\RequirePackage{fix-cm}
\documentclass[smallcondensed]{svjour3}     
\smartqed  
\usepackage{graphicx}
\usepackage{algorithm}
\usepackage[noend]{algpseudocode}
\usepackage{url}
\usepackage{amsmath} 
\usepackage{amssymb}
\usepackage{verbatim} 
\usepackage{amsfonts}
\usepackage{longtable}
\usepackage{graphicx}
\usepackage{booktabs}
\usepackage{url}
\usepackage{epstopdf}
\usepackage{longtable}
\usepackage{multirow}
\usepackage{natbib} 
\usepackage{varwidth}
\usepackage{tabularx} 
\usepackage{enumitem}

\usepackage{color}
\usepackage{xcolor}
\usepackage{pgfpages}
\usepackage{pstricks,pst-node,pst-text,pst-3d}
\usepackage{multirow}
\usepackage{multirow}
\usepackage{pstricks,pst-node,pst-text,pst-3d}
\usepackage{graphicx}
\usepackage{verbatim}
\usepackage{pifont}

\xdefinecolor{darkgreen}{rgb}{0,0.35,0}

\xdefinecolor{mygreen}{rgb}{0,0.4,0}
\xdefinecolor{myred}{rgb}{0.8,0,0}

\usepackage[titletoc,toc,title]{appendix}

\usepackage[subrefformat=parens]{subfig}

  {\arraycolsep 0pt \begin{array}{l}}%
  {\end{array}}
\algnotext{EndFor}
\algnotext{EndIf}

\newcommand{\naf}{NaF } 
\newcommand{\xhail}{XHAIL }

\newcommand{\myhappens}{\textsf{happensAt}}
\newcommand{\myholdsAt}{\textsf{holdsAt}}
\newcommand{\myinitiatedAt}{\textsf{initiatedAt}}
\newcommand{\myterminatedAt}{\textsf{terminatedAt}}
\newcommand{\mynbf}{\textsf{not}}

\newcounter{examplecounter}
\newenvironment{myexample}{
    \refstepcounter{examplecounter}%
  \textbf{Example \arabic{examplecounter}}%
  ~
}{%
\hfill $\square$

}

\newenvironment{myproof}{
    \textit{\textbf{Proof}}%
  ~
}{%
\hfill $\blacksquare$

}

\begin{document}

\title{Incremental Learning of Event Definitions with Inductive Logic Programming}

\titlerunning{Incremental Learning of Event Definitions}        

\author{Nikos Katzouris \and Alexander Artikis \and George Paliouras}


\institute{Nikos Katzouris \at
              National Center for Scientific Research ``Demokritos'' and
              National University of Athens  \\
              \email{nkatz@iit.demokritos.gr}           
           \and
              Alexander Artikis \at
              University of Piraeus and National Center for Scientific  Research ``Demokritos'' \\
           \email{\email{a.artikis@iit.demokritos.gr}}   
           \and           
           George Paliouras \at
              National Center for Scientific Research ``Demokritos'' \\
              \email{paliourg@iit.demokritos.gr}
}

\date{Received: date / Accepted: date}

\maketitle

\begin{abstract}
Event recognition systems rely on properly engineered knowledge bases of event definitions to infer occurrences of events in time. The manual development of such knowledge is a tedious and error-prone task, thus event-based applications may benefit from automated knowledge construction techniques, 
such as Inductive Logic Programming (ILP), which combines machine learning with the declarative and formal semantics of First-Order Logic. However, learning  temporal logical formalisms, which are typically utilized by logic-based Event Recognition systems is a challenging task, which most ILP systems cannot fully undertake. In addition, event-based data is usually massive and collected at different times and under various circumstances. Ideally, systems that learn from temporal data should be able to operate in an incremental mode, that is, revise prior constructed knowledge in the face of new evidence. Most ILP systems are batch learners, in the sense that in order to account for new evidence they have no alternative but to forget past knowledge and learn from scratch. Given the increased inherent complexity of ILP and the volumes of real-life temporal data, this results to algorithms that scale poorly. In this work we present an incremental method for learning and revising event-based knowledge, in the form of Event Calculus programs. The proposed algorithm relies on abductive-inductive learning and comprises a scalable clause refinement methodology, based on a compressive summarization of clause coverage in a stream of examples. We present an empirical evaluation of our approach on real and synthetic data from activity recognition and city transport applications. 
\end{abstract}

\section{Introduction}
\label{intro}

The growing amounts of temporal data collected during the execution of various tasks within organizations are hard to utilize without the assistance of  automated processes. Event Recognition ~\citep{etzion2010event,power_of_events,luckham_event_glossary} refers to the automatic detection of event occurrences within a system. From a sequence of \emph{low-level events} (for example sensor data) an event recognition system recognizes \emph{high-level events} of interest, that is, events that satisfy some pattern. Event recognition systems with a logic-based representation of event definitions, such as the Event Calculus~\citep{event_calculus}, are attracting significant attention in the event processing community for a number of reasons, including the expressiveness and understandability of the formalized knowledge, their declarative, formal semantics ~\citep{paschke,artikis2012logic} and their ability to handle rich background knowledge. Using logic programs in particular, has an extra advantage, due to the close connection between logic programming and machine learning in the field of Inductive Logic Programming (ILP)~\citep{ilp_theory_and_methods,ilp_book}. However, such applications impose challenges that make most ILP systems inappropriate. 

Several logical formalisms which incorporate time and change employ non-monotonic operators as a means for representing  commonsense phenomena \linebreak \citep{commonsense_reasoning}. Normal logic programs 
with Negation as Failure (NaF) 
in particular are a prominent non-monotonic formalism. Most ILP learners cannot handle \naf at all, or lack a robust \naf semantics~\citep{sakama_ie_naf,xhail}. Another problem that often arises when dealing with events, is the need to infer implicit or missing knowledge, for instance the indirect effects of events, or possible causes of observed events. In ILP the ability to reason with missing, or indirectly observable knowledge is called \emph{non-Observational Predicate Learning (non-OPL)} \citep{IE_and_progol}. This is a task that most ILP systems have difficulty to handle, especially when combined with \naf in the background knowledge \citep{ray_abduction_for_induction}. One way to address this problem is through the combination of ILP with Abductive Logic Programming (ALP)~\citep{alp_3,alp_1,alp_2}. Abduction in logic programming is usually given a non-monotonic semantics \citep{naf_abduction} and in addition, it is by nature an appropriate framework for reasoning with incomplete knowledge. Although it has a long history in the literature~\citep{ailp}, only recently has this combination brought about systems such as \xhail~\citep{xhail}, TAL~\citep{tal} 
and ASPAL \citep{corapi_asp_ilp,raspal} that may be used for the induction of event-based knowledge.

The above three systems which, to the best of our knowledge, are the only ILP learners that address the aforementioned learnability issues, are \emph{batch learners}, in the sense that all training data must be in place prior to the initiation of the learning process. This is not always suitable for event-oriented learning tasks, where data is often collected at different times and under various circumstances, or arrives in streams. In order to account for new training examples, a batch learner has no alternative but to re-learn a hypothesis from scratch. The cost is poor scalability when ``learning in the large'' \citep{next_ten_years} from a growing set of data. This is particularly true in the case of temporal data, which usually come in large volumes. Consider for instance data which span a large period of time, or sensor data transmitted at a very high frequency.

An alternative approach is learning incrementally, 
that is, processing training instances when they become available, and altering previously inferred knowledge to fit new observations, instead of discarding it and starting from scratch. This process, also known as \emph{Theory Revision} \citep{wrobel}, exploits previous computations to speed-up the learning, since revising a hypothesis is generally considered more efficient than learning it from scratch \citep{inthelex_database,inthelex,cattafi2010incremental}. Numerous theory revision systems have been proposed in the literature -- see \citep{inthelex} for a review --- however their applicability in non-monotonic domains is limited \citep{learning_rules_from_user_behaviour}. This issue is addressed by recent approaches to \emph{theory revision as non-monotonic ILP} \citep{learning_rules_from_user_behaviour,maggi2011revising,corapi2011norm}, where a non-monotonic learner is used to extract a set of prescriptions, which can in turn be interpreted into a set of syntactic transformations on the theory at hand. However, scaling to the large volumes of today's datasets or handling streaming data remains an open issue, and the development of scalable algorithms for theory revision has been identified as an important research direction \citep{ilp_turns_20}. As historical data grow over time, it becomes progressively harder to revise knowledge, so that it accounts both for  new evidence and past experience. One direction towards scaling theory revision systems is the development of techniques for reducing the need for reconsulting  the whole  history of accumulated experience, while updating existing knowledge.  

This is the direction we take in this work. We build on the ideas of non-monotonic ILP and use \xhail as the basis for a scalable, incremental learner for the induction of event definitions in the form of  Event Calculus theories. XHAIL has been used for the induction of action theories~\citep{engineering_policybased_ubiquitous_systems,deriving_nonzeno_behaviour_models_from_goal_models_using_ILP,an_inductive_approach_for_modal_transition_system_refinement,learning_from_vacuously_satisfiable_scenariobased_specifications,learning_operational_requirements_from_goal_models}. Moreover, in~\citep{learning_rules_from_user_behaviour} it has been used for theory revision in an incremental setting, revising hypotheses with respect to a recent, user-defined subset of the perceived experience. In contrast, the learner we present here performs revisions that account for all examples seen so far.  We describe a 
compressive ``memory'' structure, incorporated in the learning process, which reduces the need for reconsulting past experience in response to a revision. Using this structure, we propose a method which, given a stream of examples, a theory which accounts for them and a new training instance, requires at most one pass over the examples in order to revise the initial theory, so that it accounts for both past and new evidence.
We evaluate empirically our approach on real and synthetic data from an activity recognition application and a transport management application. Our results indicate that our approach is significantly more efficient than XHAIL, without compromising predictive accuracy, and scales adequately to large data volumes.     

The rest of this paper is structured as follows. Section \ref{sec:background} provides a brief overview of abductive and inductive logic programming. In section \ref{sec:event_related} we present the Event Calculus dialect that we employ, describe the domain of activity recognition that we use as a running example and show how event definitions may be learnt using XHAIL. In Section \ref{sec:incremental_learning} we present our proposed method, prove its correctness and present the details of its abductive-inductive mechanism. In Section \ref{sec:discussion} we discuss some theoretical and practical implications of our approach. In Section \ref{sec:experiments} we present the experimental evaluation, and finally in Sections \ref{sec:related} and \ref{sec:concl} we discuss related work and draw our main conclusions.

\section{Background}
\label{sec:background}

We assume a first-order language as in \citep{lp_foundations_lloyd} where \mynbf \ in front of literals denotes Negation as Failure (NaF). We call a logic program \emph{Horn} if it is NaF-free and \emph{normal} otherwise. For more details on the basic terminology and conventions of logic programming used in this work see Appendix \ref{app:appendix_lp}. We define the entailment relation between normal logic programs in terms of the \emph{stable model semantics} \citep{stable_model_semantics} and in particular its credulous version, under which program $\Pi_1$ entails program $\Pi_2$, denoted by $\Pi_1 \vDash \Pi_2$, if at least one stable model of $\Pi_1$ is a stable model of $\Pi_2$. Following Prolog's convention, throughout this paper, predicates and ground terms in logical formulae start with a lower case letter, while variable terms start with a capital letter. 

Inductive Logic Programming (ILP) is a subfield of machine learning based on logic programming. Given a set of positive and negative examples represented as logical facts, an ILP algorithm derives a set of non-ground rules which discriminate between the positive and the negative examples, potentially taking into account some background knowledge. Definition \ref{def:ilp} provides a formal account. 

\begin{definition}[ILP]
\label{def:ilp}
An ILP task is a triplet $ILP(B,E,M)$ where $B$ is a normal logic program, $E = E^{+}\cup E^{-}$ is a set of ground literals called positive ($E^{+}$) and negative ($E^{-}$) examples and $M$ is a set of clauses called \emph{language bias}. A normal logic program $H$ is called an inductive hypothesis for the ILP task if $H\subseteq M$ and $B\cup H$ \emph{covers the examples}, that is, $B\cup H\vDash E^{+}$ and $B\cup H\nvDash E^{-}$. 
\end{definition}

The language bias mentioned in Definition \ref{def:ilp} reduces the complexity of an ILP task by imposing syntactic restrictions on hypotheses that may be learnt. A commonly used language bias in ILP, also employed in this work is \emph{mode declarations} \citep{IE_and_progol}. A mode declaration is a template literal that can be placed either in the head or the body of a hypothesis clause and contains special placemarkers for variables and ground terms. A set of mode declarations $M$ defines a language $\mathcal{L}(M)$, called \emph{mode language}. A clause is in $\mathcal{L}(M)$ if it is constructed from head and body mode declarations by replacing variable placemarkers by actual variable symbols and ground placemarkers by ground terms. Formal definitions for mode declarations and the mode language are provided in Appendix \ref{app:appendix_lp}. 
ILP algorithms that use mode declarations work by using $\mathcal{L}(M)$ as a search space for clauses, trying to optimize an objective function which takes into account example coverage and hypothesis size. Typically, the search space $\mathcal{L}(M)$ is structured via $\theta$-subsumption. 

\begin{definition}[$\mathbf{\theta}$-\textbf{subsumption}]
Clause $C$ $\theta$-subsumes clause $D$, denoted $C \preceq D$, if there exists a substitution $\theta$ such that $head(C)\theta = \ head(D)$ and $body(C)\theta \subseteq body(D)$, where $\mathit{head(C)}$ and $\mathit{body(C)}$ denote the head and the body of clause $C$ respectively. Program $\Pi_1$ $\theta$-subsumes program $\Pi_2$ if for each clause $C\in \Pi_1$ there exists a clause $D\in \Pi_2$ such that $C\preceq D$.
\end{definition}

$\theta$-subsumption provides a syntactic notion of generality \citep{dvzeroski2010relational} which may be used to search for clauses based on their example coverage. Clause $C$ is more general than clause $D$ (resp. $D$ is more specific than $C$) if $C\preceq D$, in which case the examples covered by $D$ are a subset of the examples covered by $C$. The generality order between clauses is naturally extended to hypotheses via $\theta$-subsumption between programs.

Given an ILP task $\mathit{ILP(B,E,M)}$, a hypothesis $H$ is called \emph{incomplete} if $B\cup H$ does not cover some positive examples from $E$ and \emph{inconsistent} if it covers some negative examples. An inductive hypothesis for the ILP task, that is, a hypothesis that is both complete and consistent, is called \emph{correct}. An incomplete hypothesis $H$ can be made complete by \emph{generalization}, that is, a set of syntactic transformations that aim to increase example coverage, and may include the addition of new clauses, or the removal of literals from existing clauses. Similarly, an inconsistent hypothesis can be made consistent by \emph{specialization}, a process that aims to restrict example coverage and may include removal of clauses from the hypothesis, or addition of new literals to existing clauses in the hypothesis. Theory revision is the process of acting upon a hypothesis by means of syntactic transformations (\emph{generalization and specialization operators}), in order to change the answer set of the hypothesis \citep{wrobel,inthelex}, that is, the examples it accounts for. Theory revision is at the core of incremental ILP systems. In an incremental setting, examples are provided over time. A learner induces a hypothesis from scratch, from the first available set of examples, and treats this hypothesis as a revisable background theory in order to account for new examples.

\section{Event Calculus and Machine Learning for Event Recognition}
\label{sec:event_related}

\scriptsize

\begin{table}[t]
\begin{center}
\renewcommand{\arraystretch}{0.9}
\setlength\tabcolsep{3pt}
\begin{tabular}{ll}
\hline\noalign{\smallskip}
\multicolumn{1}{c}{\textbf{Predicate}} & \multicolumn{1}{c}{\textbf{Meaning}} \\
\noalign{\smallskip}
\hline
\noalign{\smallskip}
\myhappens$(E,T)$ & Event $E$ occurs at time $T$ \\[8pt]

\myinitiatedAt$(F,T)$ & At time $T$ a period of time  \\
&  for which fluent $F$ holds is initiated \\[8pt]

\myterminatedAt$(F,T)$ & At time $T$ a period of time \\
&  for which fluent $F$ holds is terminated \\[8pt]


\myholdsAt$(F,T)$ & Fluent $F$ holds at time $T$ \\[8pt]
\hline
\noalign{\smallskip}
\multicolumn{1}{c}{\textbf{Axioms}} & \multicolumn{1}{c}{~} \\
\noalign{\smallskip}
\hline
$\begin{array}{lr}
 \myholdsAt(F,T+1) \leftarrow \\
  \qquad  \myinitiatedAt(F,T).\\
\end{array}$
&
 \  \  \ \ \  \ \ \ \ $\begin{array}{lr}
     \\
     \myholdsAt(F,T+1) \leftarrow \\
     \qquad \myholdsAt(F,T),\\
     \qquad \mynbf \  \myterminatedAt(F,T).  \\
    \end{array}$ \\
\noalign{\smallskip}
\hline
\end{tabular}
\caption{\small The basic predicates and axioms of \textsf{SDEC}}
\label{table:ec}
\end{center}

\end{table}
\normalsize

The Event Calculus \citep{event_calculus} is a temporal logic for reasoning about events and their effects. It is a formalism that has been successfully used in numerous event recognition applications \citep{paschke,artikisTKDE,chaudet2006extending,cervesato2000calculus}. The ontology of the Event Calculus comprises \emph{time points}, i.e. integers of real numbers; \emph{fluents}, i.e. properties which have certain values in time; and \emph{events}, i.e. occurrences in time that may affect fluents and alter their value. The domain-independent axioms of the formalism incorporate the common sense \emph{law of inertia}, according to which fluents persist over time, unless they are affected by an event.
We call the Event Calculus dialect used in this work Simplified Discrete Event Calculus (\textsf{SDEC}). As its name implies, it is a simplified version of the Discrete Event Calculus, a dialect which is equivalent to the classical Event Calculus when time ranges over integer domains \citep{mueller_ec}. 

The building blocks of \textsf{SDEC} and its domain-independent axioms are presented in Table \ref{table:ec}. The first axiom in Table \ref{table:ec} states that a fluent $F$ holds at time $T$ if it has been initiated at the previous time point, while the second axiom states that $F$ continues to hold unless it is terminated. \myinitiatedAt/2 ~  and \myterminatedAt/2 ~ are defined in an application-specific manner. Examples will be presented shortly.

\subsection{Running example: Activity recognition}
\label{sec:activity_recognition}

\scriptsize

\begin{table}[t]
\small
\begin{center}
\renewcommand{\arraystretch}{0.9}
\setlength\tabcolsep{3pt}
\begin{tabular}{ll}
\hline\noalign{\smallskip}
\multicolumn{1}{c}{\textbf{Narrative}} & \multicolumn{1}{c}{\textbf{Annotation}}  \\
\noalign{\smallskip}
\hline
\noalign{\smallskip}
...... & ...... \\
\noalign{\smallskip}
 \myhappens$(inactive(id_1),999)$        & \mynbf \  \myholdsAt$(moving(id_1,id_2),999)$  \\
 \myhappens$(active(id_2),999)$          & ~  \\
 \myholdsAt$(coords(id_1,201,432),999)$  &  ~  \\
 \myholdsAt$(coords(id_2,230,460),999)$  &    ~   \\
 \myholdsAt$(direction(id_1,270),999)$   & ~  \\
 \myholdsAt$(direction(id_2,270),999)$   & ~  \\
                          
~ & ~\\

 \myhappens$(walking(id_1),1000)$          & \mynbf \  \myholdsAt$(moving(id_1,id_2),1000)$  \\
 \myhappens$(walking(id_2),1000)$          & ~  \\
 \myholdsAt$(coords(id_1,201,454),1000)$  &  ~  \\
 \myholdsAt$(coords(id_2,230,440),1000)$  &    ~   \\
 \myholdsAt$(direction(id_1,270),1000)$   & ~  \\
 \myholdsAt$(direction(id_2,270),1000)$   & ~  \\

~ & ~\\

 \myhappens$(walking(id_1),1001)$          & \   \myholdsAt$(moving(id_1,id_2),1001)$  \\
 \myhappens$(walking(id_2),1001)$          & ~  \\
 \myholdsAt$(coords(id_1,201,454),1001)$  &   ~  \\
 \myholdsAt$(coords(id_2,227,440),1001)$  &   \  \  ~   \\
 \myholdsAt$(direction(id_1,275),1001)$   & ~  \\
 \myholdsAt$(direction(id_2,278),1001)$   & ~  \\

 ...... & ......\\

\hline
\end{tabular}
\caption{\small An annotated stream of LLEs}\label{table:stream1}
\end{center}
\end{table}
\normalsize

Throughout this paper we use the task of activity recognition, as defined in the CAVIAR\footnote{http://homepages.inf.ed.ac.uk/rbf/CAVIARDATA1/} project, as a running example. The CAVIAR dataset consists of videos of a public space, where actors walk around, meet each other, browse information displays, fight and so on. These videos have been manually annotated by the CAVIAR team to provide the ground truth for two types of activity. The first type corresponds to low-level events, that is, knowledge about a person's activities at a certain time point (for instance \emph{walking}, \emph{running}, \emph{standing still} and so on). The second type corresponds to high-level events,  activities that involve more than one person, for instance two people \emph{moving together}, \emph{
fighting}, \emph{meeting} and so on. The aim is to recognize high-level events by means of combinations of low-level events and some additional domain knowledge, such as a person's position and direction at a certain time point. 

Low-level events are represented in \textsf{SDEC} by streams of ground \myhappens/2 atoms (see Table \ref{table:stream1}), while high-level events and other domain knowledge are represented by ground \myholdsAt/2 atoms. Streams of low-level events together with domain-specific knowledge will henceforth constitute the \emph{narrative}, in ILP terminology, while knowledge about high-level events is the \emph{annotation}. Table \ref{table:stream1} presents an annotated stream of low-level events. We can see for instance that the person $id_1$ is $inactive$ at time $999$, her $(x,y)$ coordinates are $(201,432)$ and her direction is $270^{\circ}$. The annotation for the same time point informs us that $id_1$ and $id_2$ are not moving together. Fluents express both high-level events and input information, such as the coordinates of a person. We discriminate between \emph{inertial} and \emph{statically defined} fluents. The former should be inferred by the Event Calculus axioms, while the latter are provided with the input. 

Given such a domain description in the language of \textsf{SDEC}, the aim of machine learning addressed in this work is to automatically derive the \emph{Domain-Specific Axioms}, that is, the axioms that specify how the occurrence of low-level events affects the truth values of the fluents that represent high-level events, by initiating or terminating them. Thus, we wish to learn \myinitiatedAt/2 and \myterminatedAt/2 definitions from positive and negative examples from the narrative and the annotation. 

Henceforth, we use the term ``example'' to encompass  anything known true at a specific time point. We assume a closed world, thus anything that is not explicitly given is considered false (to avoid confusion, in the tables throughout the paper we state both negative and positive examples). An example's time point will also serve as reference. For instance, three different examples $e_{999}, e_{1000}$ and $e_{1001}$ are presented in Table \ref{table:stream1}. According to the annotation, an example is either positive or negative w.r.t. a particular high-level event. For instance, $e_{1000}$ in Table \ref{table:stream1} is a negative example for the \emph{moving} high-level event, while $e_{1001}$ is a positive example.

\subsection{Learning and Revising Event Definitions}
\label{sec:learning_difficulties}

Learning event definitions in the form of domain-specific Event Calculus axioms with ILP poses several challenges. Note first, that the learning problem presented in Section \ref{sec:activity_recognition} requires non-Observational Predicate Learning (non-OPL) \citep{IE_and_progol}, meaning that instances of target predicates (\textsf{initiatedAt/2} and \textsf{terminatedAt/2}) are not provided with the supervision. Using abduction to obtain the missing instances is a solution. Abduction is a form of logical inference that seeks to extract a set of explanations that make a set of observations true. In Abductive Logic Programming (ALP) the observations are represented by a set of queries, and one derives explanations for these observations in the form of ground facts that make the queries succeed. Definition \ref{def:alp} provides a formal account.

\begin{definition}[ALP]
\label{def:alp}
An ALP task is a triplet $ALP(B,A,G)$ where $B$ is a normal logic program, $A$ is a set of predicates called \emph{abducibles} and $G$ is a set of ground queries called \emph{goals}. A set of ground atoms $\Delta$ is called an \emph{abductive explanation} for the ALP task if the predicate of each atom in $\Delta$ appears in $A$ and $B\cup \Delta \vDash G$.
\end{definition} 

Using ALP, the missing supervision for the learning problem of Section \ref{sec:activity_recognition} can be obtained by abducing a set of ground \myinitiatedAt/2 and \myterminatedAt/2 atoms as explanations for the conjunction of the \myholdsAt/2 literals of the annotation (see Table \ref{table:stream1}). In principle, several explanations are possible for a given set of observations. To avoid redundant explanations, ALP reasoners are typically biased towards \emph{minimal} explanations. For instance, the atom $\mathit{\myinitiatedAt(moving(id_1,id_2),1000)}$ is a minimal abductive explanation for the \myholdsAt/2 literals in Table \ref{table:stream1}.

Several systems have been proposed that combine ILP with abductive reasoning. These systems use abduction to obtain missing knowledge, necessary to explain the provided examples, and then employ standard ILP techniques to construct hypotheses. However most of these systems cannot be used for learning Event Calculus programs. Some of these abductive-inductive systems are restricted to Horn logic (HAIL \citep{hail}, IMPARO \citep{imparo}). Others can handle negation, but their use of abduction is limited. For instance INTHELEX \citep{inthelex} uses abduction only to generate facts that might be missing from the description of an example, and is otherwise restricted to OPL. PROGOL5 \citep{Muggleton_theory_compl}, ALEPH\footnote{\url{http://web.comlab.ox.ac.uk/oucl/research/areas/machlearn/Aleph/}} and ALECTO \citep{alecto} support some form of abductive reasoning but lack the full power of ALP. As a result, they cannot reason abductively with negated atoms \citep{ray_abduction_for_induction}. 

NaF is responsible for two more shortcomings of traditional ILP approaches w.r.t. normal logic programs. First, as explained in \citep{ray_abduction_for_induction}, the standard set cover approach on which most ILP systems rely, is essentially unsound in the presence of NaF, meaning that it may return hypotheses that do not cover all the examples. Because of NaF and its non-monotonicity, newly inferred clauses may be invalidated by past examples. At the same time the learner has no way to detect that, because in a set cover approach, designed to operate under the monotonicity of Horn logic, past examples are retracted from memory once they are covered by a clause.

The second shortcoming concerns theory revision, and is related to the standard $\theta$-subsumption-based heuristics used in Horn logic, which are known to be inapplicable in general in the case of normal logic programs \citep{fogel1998normal}. ILP systems construct clauses either in a bottom-up, or a top-down manner, i.e. searching for more general or more specific hypotheses respectively, in a space ordered by $\theta$-subsumption. This is an acceptable strategy to guide the search in Horn logic, because in this case, ``moving up'' the subsumption lattice, i.e. from specific to general, increases example coverage, while ``moving down'', from general to specific, restricts example coverage. This does not always hold in normal logic programs, where generalizing (resp. specializing) a single clause in a hypothesis may result in less (resp. more) examples covered by the hypothesis. As a result, revising a hypothesis in a clause-by-clause manner using subsumption to guide the search, cannot be used in full clausal logic. We illustrate the case with a simple example.

\begin{myexample}
\label{ex:terminated_non_monotonic}
Consider the following annotated narrative related to the \emph{fighting} high-level event from CAVIAR: 

\small
\begin{equation*}
\label{eq:no_locality_example}
\begin{array}{lr}
\mathbf{Narrative:}\\
\myhappens(abrupt(id_1),1).\\
\myhappens(abrupt(id_2),1).\\
\myholdsAt(close(id_1,id_2,23),1).\\
\myhappens(walking(id_1),2).\\
\myhappens(abrupt(id_2),2).\\
\myholdsAt(close(id_1,id_2,23),2).\\
\end{array}
\quad
\begin{array}{lr}
\mathbf{Annotation:}\\

\mynbf \ \myholdsAt(fighting(id_1,id_2),1).\\
\myholdsAt(fighting(id_1,id_2),2).\\
\myholdsAt(fighting(id_1,id_2),3).\\
\\
\\ \\ 
\end{array}
\end{equation*}
\normalsize

\noindent where $close(X,Y,D)$ is a statically defined fluent which states that the Euclidean distance  between persons $X$ and $Y$ is less than threshold $D$. Consider also the clauses: 

\small
\noindent\begin{tabularx}{\textwidth}{@{}XXX@{}}
  \begin{equation*}
    \label{eq:ec_axioms_1}
    \begin{array}{lr}
    C_1 = \myinitiatedAt(fighting(X,Y),T) \leftarrow \\
    \qquad \qquad \myhappens(abrupt(X),T),\\
    \qquad \qquad \mynbf \ \myhappens(inactive(Y),T),\\
    \qquad \qquad \myholdsAt(close(X,Y,23),T).
    \end{array}
  \end{equation*} & 
   \begin{equation*}
    \label{eq:ec_axioms_2} 
    \begin{array}{lr}
    C_2 = \myterminatedAt(fighting(X,Y),T) \leftarrow \\
    \qquad \qquad \myhappens(walking(X),T).\\
    \end{array}
  \end{equation*}
\end{tabularx}  

\vspace*{-0.5cm}

\noindent\begin{tabularx}{\textwidth}{@{}XXX@{}}
  \begin{equation*}
    \label{eq:ec_axioms_3}
    \begin{array}{lr}
    C_{2}' = \myterminatedAt(fighting(X,Y),T) \leftarrow \\
\qquad \qquad \myhappens(walking(X),T),\\
\qquad \qquad \mynbf \ \myholdsAt(close(X,Y,23),T).\\
    \end{array}
  \end{equation*} & 
 \end{tabularx} 
\normalsize

\noindent Clause $C_1$ states that \emph{fighting} between two persons $id_1$  and $id_2$ is initiated if one of them exhibits an \emph{abrupt} behavior, the other is not \emph{inactive} and their distance is less than 23 pixel positions on the video frame. Clause $C_2$ states that \emph{fighting} is terminated between two people if one of them walks. Clause $C_{2}'$ is a specialization of $C_2$ and dictates that \emph{fighting} between two persons is terminated when one of them walks away. Consider two hypotheses $H_1,H_2$ where $H_1=\{C_1,C_2\}$ and $H_2=\{C_1,C_{2}'\}$. Observe that $\mathsf{SDEC} \cup H_1$ is an incomplete hypothesis, because it does not cover the positive example $\myholdsAt(fighting(id_1,id_2),3)$. Indeed, by means of clause $C_2$ the fluent $\mathit{fighting(id_1,id_2)}$ is terminated at time 2, and thus it does not hold at time 3. On the other hand, hypothesis $\mathsf{SDEC} \cup H_2$ does cover the positive example at time 3 because clause $C_{2}'$ does not terminate the \emph{fighting} fluent at time 2. We thus have that hypothesis $H_2$, though more specific than $H_1$, covers more examples. 
\end{myexample}

Recently, a number of hybrid ILP-ALP systems  have been proposed, that are able to overcome the aforementioned shortcomings. XHAIL is one such system, which is at the basis of our approach to learning event definitions from streams of event-based knowledge. We next give a detailed account of XHAIL.

\vspace*{-6cm}

\subsubsection{The XHAIL System}
\label{sec:xhail}

\vspace*{-0.5cm}

\xhail constructs hypotheses in a three-phase process. Given an ILP task \linebreak $\mathit{ILP(B,E,M)}$, the first two phases return a ground program $K$, called \emph{Kernel Set of $E$}, such that $B\cup K\vDash E$. The first phase generates the heads of $K$'s clauses by abductively deriving from $B$ a set $\Delta$ of instances of head mode declaration atoms, such that $B\cup \Delta \vDash E$. The second phase generates $K$, by saturating each previously abduced atom with instances of body declaration atoms that deductively follow from $B\cup \Delta$.

\begin{table}[H]
\scriptsize
\begin{center}
\renewcommand{\arraystretch}{0.9}
\setlength{\tabcolsep}{2pt}
\scalebox{0.9}{
\begin{tabular}{ll}
\noalign{\smallskip}
\hline
\noalign{\smallskip}
\multicolumn{1}{l}{\textbf{Input}} & \multicolumn{1}{l}{}  \\
\hline\noalign{\smallskip}
~ & ~ \\
\textbf{Narrative} & \textbf{Annotation}  \\
~ & ~ \\
$\mathit{\myhappens(abrupt(id_1), \ 1).}$ & $\mathit{\myholdsAt(fighting(id_1,id_2), \ 1).}$\\
$\mathit{\myhappens(walking(id_2), \ 1).}$ & $\mathit{\mynbf \ \myholdsAt(fighting(id_3,id_4), \ 1).}$   \\
$\mathit{\mynbf \ \myholdsAt(close(id_1,id_2,23), \ 1).}$ & $\mathit{\mynbf \ \myholdsAt(fighting(id_1,id_2), \ 2).}$  \\
$\mathit{\myhappens(abrupt(id_3), \ 2).}$ & $\mathit{\mynbf \ \myholdsAt(fighting(id_3,id_4), \ 2).}$ \\
$\mathit{\myhappens(abrupt(id_4), \ 2).}$ & $\mathit{\mynbf \ \myholdsAt(fighting(id_1,id_2), \ 3).}$    \\
$\mathit{\myholdsAt(close(id_3,id_4,23), \ 2).}$ & $\mathit{\myholdsAt(fighting(id_3,id_4), \ 3).}$  \\
~ & ~ \\
\textbf{Mode declarations} & \textbf{Background knowledge}  \\
~ & ~ \\
$\mathit{modeh(\myinitiatedAt(moving(+pid,+pid), \ +time))}$ &  Axioms of \textsf{SDEC} (Table \ref{table:ec})\\
$\mathit{modeh(\myterminatedAt(moving(+pid,+pid), \ +time))}$ & \\
$ \mathit{modeb(\myhappens(walking(+pid), \ +time))}$ & \\
$\mathit{modeb(\myhappens(abrupt(+pid), \ +time))}$ & \\
$\mathit{modeb(\myholdsAt(close(+pid,+pid,\#dist), \ +time))}$ &\\

~ & ~ \\
\hline\noalign{\smallskip}
\textbf{Phase 1 (Abduction):} & ~ \\
\hline\noalign{\smallskip}
~ & ~ \\
$\mathit{\Delta_1 = \{ \myinitiatedAt(fighting(id_3,id_4), \ 2)},$& \\
 \qquad \quad $\mathit{\myterminatedAt(fighting(id_1,id_2), \ 1) \}}$ & \\
~ & ~ \\

\hline\noalign{\smallskip}
\textbf{Phase 2 (Deduction):} & ~ \\
\hline\noalign{\smallskip}
~ & ~ \\
\textbf{Kernel Set} $K$: & \textbf{Variabilized Kernel Set} $K_v$: \\
~ & ~ \\
 $\mathit{\myinitiatedAt(fighting(id_3,id_4), \ 2)}\  \leftarrow$ & $\mathit{\myinitiatedAt(fighting(X,Y), \ T)}\  \leftarrow$ \\
 $\quad \mathit{\myhappens(abrupt(id_3), \ 2),} $ & $\quad \mathit{\myhappens(abrupt(X), \ T),}$\\
 $ \quad \mathit{\myhappens(abrupt(id_4), \ 2),}$ & $\quad \mathit{\myhappens(abrupt(Y), \ T),}$\\
 $\quad \mathit{\myholdsAt(close(id_3,id_4,23), \ 2).}$ & $\quad \mathit{\myholdsAt(close(X,Y,23), \ T).}$\\
~ & ~\\
 $\mathit{\myterminatedAt(fighting(id_1,id_2), \ 1)}\  \leftarrow$ & $\mathit{\myterminatedAt(fighting(X,Y), \ T)}\  \leftarrow$ \\
 $\quad \mathit{\myhappens(abrupt(id_1), \ 1),}$ & $\quad \mathit{\myhappens(abrupt(X), \ T),}$\\
 $\quad \mathit{\myhappens(walking(id_2), \ 1),}$ & $\quad \mathit{\myhappens(walking(Y), \ T),}$\\
 $\quad \mathit{\mynbf \ \myholdsAt(close(id_1,id_2,23), \ 1).}$ & $\quad \mathit{\mynbf \ \myholdsAt(close(X,Y,23), \ T).}$\\

~ & ~ \\

\hline\noalign{\smallskip}
\multicolumn{2}{l}{\textbf{Phase 3 (Induction):}}  \\
\hline\noalign{\smallskip}
~ & ~ \\
\textbf{Program} $U_{K_v}$ (\textbf{Syntactic transformation of} $K_v$):\\
~ & ~ \\
$\mathit{\myinitiatedAt(fighting(X,Y), \ T)}\  \leftarrow$ & $\mathit{\myterminatedAt(fighting(X,Y), \ T)}\  \leftarrow$ \\
$\quad \mathit{use(1,0),try(1,1,v(X,T)),}$ & $\quad \mathit{use(2,0),try(2,1,v(X,T)),}$\\
$\quad \mathit{try(1,2,v(Y,T)),}$ & $\quad \mathit{try(2,2,v(Y,T)),}$\\
$\quad \mathit{try(1,3,v(X,Y,T)).}$ & $\quad \mathit{try(2,3,v(X,Y,T)).}$\\ 
~ & ~ \\   
$\mathit{try(1,1,v(X,T))}\  \leftarrow$ & $\mathit{try(2,1,v(X,T))}\  \leftarrow $ \\
$\quad \mathit{use(1,1),\myhappens(abrupt(X), \ T).}$ & $\quad \mathit{use(2,1),\myhappens(abrupt(X), \ T).}$\\
$\mathit{try(1,1,v(X,T))}\  \leftarrow \mynbf \ use(1,1).$ & $\mathit{try(2,1,v(X,T))}\  \leftarrow \mynbf \ use(1,1).$ \\
~ & ~ \\ 
$\mathit{try(1,2,v(Y,T))}\  \leftarrow$ & $\mathit{try(2,2,v(Y,T))}\  \leftarrow$ \\
$\quad \mathit{use(1,2),\myhappens(abrupt(Y), \ T).}$ & $\quad \mathit{use(2,2),\myhappens(walking(Y), \ T).}$\\
$\mathit{try(1,2,v(X,T))}\  \leftarrow \mynbf \ use(1,2).$ & $\mathit{try(2,2,v(Y,T))}\  \leftarrow \mynbf \ use(2,2).$ \\
~ & ~ \\    
$\mathit{try(1,3,v(X,Y,T))}\  \leftarrow$ & $\mathit{try(2,3,v(X,Y,T))}\  \leftarrow$ \\
$\quad \mathit{use(1,3),\myholdsAt(close(X,Y,23), \ T).}$ & $\quad \mathit{use(2,3),\mynbf \ \myholdsAt(close(X,Y,23), \ T).}$\\
$\mathit{try(1,3,v(X,T))}\  \leftarrow \mynbf \ use(1,3).$ & $\mathit{try(2,3,v(X,Y,T))}\  \leftarrow \mynbf \ use(2,3).$ \\
~ & ~ \\ 
\textbf{Search:} & \textbf{Abductive Solution:} \\
~ & ~ \\ 
$\mathit{ALP(\mathsf{SDEC}\cup U_{K_v},\{use/2\},Narrative\cup Annotation})$ & $\Delta_2 = \mathit{\{use(1,0),use(1,3),}$\\
~ & $\qquad \quad \mathit{use(2,0),use(2,2)}\}$ \\
~ & ~ \\ 

\hline\noalign{\smallskip}
\multicolumn{1}{l}{\textbf{Output hypothesis}} & \multicolumn{1}{l}{}  
\\ 
\hline\noalign{\smallskip}
~ & ~ \\
$\mathit{\myinitiatedAt(fighting(X,Y), \ T)}\  \leftarrow$ & $\mathit{\myterminatedAt(fighting(X,Y), \ T)}\  \leftarrow$ \\
$\quad \mathit{\myholdsAt(close(X,Y,23), \ T).}$ & $\quad \mathit{\myhappens(walking(Y), \ T).}$\\
~ & ~ \\
\hline
\end{tabular}
} 
\caption{\small Hypothesis generation by XHAIL. }\label{table:xhail_functionality}
\end{center}
\end{table}
\normalsize

\noindent \begin{myexample}
\label{ex:xhail_functionality_1}
Table \ref{table:xhail_functionality} presents the process of hypothesis generation by XHAIL, using an example from CAVIAR's \emph{fighting} high-level event. The input consists of examples in the from of narrative and annotation, a set of mode declarations and the axioms of \textsf{SDEC} as background knowledge. Mode declarations specify atoms that are allowed in the heads of clauses and literals that are allowed in the bodies of clauses, by being input to the $\mathit{modeh/1}$ and $\mathit{modeb/1}$ predicates respectively. Variable and ground placemarkers are indicated by terms of the form $\mathit{+type}$ and $\mathit{\#type}$ respectively. Variables in the mode declarations shown in Table \ref{table:xhail_functionality} are either of type $\mathit{pid}$, representing the $\mathit{id}$ of a person, or of type $\mathit{time}$. The only ground term that is allowed in generated literals is of type $\mathit{dist}$, representing the Euclidean distance between persons.   

The annotation says that $\mathit{fighting}$ between persons $id_1$ and $id_2$ holds at time 1 and it does not hold at times 2 and 3, hence it is terminated at time 1. Respectively, \emph{fighting} between persons $id_3$ and $id_4$ holds at time 3 and does not hold at times 1 and 2, hence it is initiated at time 2. XHAIL obtains these explanations for the \myholdsAt/2 literals of the annotation abductively, using the $\mathit{modeh}$ atoms in the mode declarations as abducible predicates. In its first phase, it derives the two ground atoms in $\Delta_1$, presented in Phase 1 of Table \ref{table:xhail_functionality}. In its second phase, XHAIL forms a Kernel Set, as presented in Phase 2 of Table \ref{table:xhail_functionality}, by generating one clause from each abduced atom in $\Delta_1$, using this atom as the head, and body literals that deductively follow from $\mathsf{SDEC}\cup \Delta_1$ as the body of the clause.
\end{myexample}

The Kernel Set is a multi-clause version of the \emph{Bottom Clause}, a concept widely used by inverse entailment systems like PROGOL and ALEPH. These systems construct hypotheses one clause at a time, using a positive example as a ``seed'', from which a most-specific Bottom Clause is generated by inverse entailment \citep{IE_and_progol}. A ``good'', in terms of some heuristic function, hypothesis clause is then constructed by a search in the space of clauses that subsume the Bottom Clause. In contrast, the Kernel Set is generated from all positive examples at once, and XHAIL performs a search in the space of theories that subsume it, in order to arrive at a ``good'' hypothesis. This is necessary due to the difficulties mentioned in Section \ref{sec:learning_difficulties}, related to the non-monotonicity of NaF, which are typical of systems that learn one clause at a time. Another important difference between the Kernel Set and the Bottom Clause is that the latter is constructed by a seed example that must be provided by the supervision, while the former can also utilize  atoms that are derived abductively from the background knowledge, allowing to successfully address non-OPL problems mentioned in Section \ref{sec:learning_difficulties}.


In order to utilize the Kernel Set as a search space, it first needs to be \emph{variabilized}. Variabilization is a process that turns each ground clause in the Kernel Set to a clause in the mode language  $\mathcal{L}(M)$ (Definition \ref{def:mode_language} of Appendix \ref{app:appendix_lp}), where $M$ denotes the input mode declarations. To do so, each term in a Kernel Set clause that corresponds to a variable, as indicated by the mode declarations, is replaced by an actual variable, while each term that corresponds to a ground term is retained intact.

\begin{myexample}[\textbf{Example \ref{ex:xhail_functionality_1} continued}]
In Table \ref{table:xhail_functionality} the variabilized Kernel Set $K_v$ is presented in Phase 2. All variable placemarkers in the mode declarations indicate input $(+)$ variables, meaning that the corresponding variable should either appear in the head of the clause, or be an output $(-)$ variable in some preceding body literal. In the absence of output variable placemarkers in the mode declarations of Table \ref{table:xhail_functionality}, each variable that appears in the body of a clause $C\in K_v$, also appears in the head of $C$. Note also that the ground term that represents a distance threshold in the $\mathit{close/3}$ predicate has been preserved during the variabilization process, since it replaces a ground placemarker in the corresponding mode declaration.
\end{myexample}

The third phase of XHAIL functionality concerns the actual search for a hypothesis. Contrary to other inverse entailment systems like PROGOL and ALEPH, which rely on a heuristic search, XHAIL performs a complete search in the space of theories that subsume $K_v$ in order to ensure soundness of the generated hypothesis. This search is biased by \emph{minimality}, i.e. preference towards hypotheses with fewer literals. A hypothesis is thus constructed  by dropping as many literals and clauses from $K_v$ as possible, while correctly accounting for all the examples. To this end, $K_v$ is subject to a syntactic transformation of its clauses, which involves two new predicates $\mathit{try/3}$ and $\mathit{use/2}$ (see Phase 3 of Table \ref{table:xhail_functionality}).  

For each clause $C_i \in K_v$ and each body literal $\delta_{i}^{j}\in C_i$, a new atom $\mathit{v(\delta_{i}^{j}})$ is generated, as a special term that contains the variables that appear in $\delta_{i}^{j}$. The new atom is wrapped inside an atom of the form $\mathit{try(i,j,v(\delta_{i}^{j}))}$. An extra atom $\mathit{use(i,0)}$ is added to the body of $C_i$ and two new clauses $\mathit{try(i,j,v(\delta_{i}^{j})) \leftarrow use(i,j), \delta_{i}^{j}}$ and $\mathit{try(i,j,v(\delta_{i}^{j})) \leftarrow \mynbf \ use(i,j)}$ are generated, for each body literal $\delta_{i}^{j} \in C_i$. All these clauses are put together into a program $U_{K_v}$ as in Table \ref{table:xhail_functionality}. $U_{K_v}$ serves as a ``defeasible'' version of $K_v$ from which literals and clauses may be selected in order to construct a hypothesis that accounts for the examples. This is realized by solving an ALP task with $\mathit{use/2}$ as the only abducible predicate, as in Phase 3 of Table \ref{table:xhail_functionality}. As explained in \citep{xhail}, the intuition is as follows: In order for the head atom of clause $C_i \in U_{K_v}$ to contribute towards the coverage of an example, each of its $\mathit{try(i,j,v(\delta_{i}^{j}))}$ atoms must succeed. By means of the two rules added for each such atom, this can be achieved in two ways: Either by assuming $\mynbf \ use(i,j)$, or by satisfying $\delta_{i}^{j}$ and abducing $use(i,j)$. A hypothesis clause is constructed by the head atom of the $i$-th clause $C_i$ of $K_v$, if  $use(i,0)$ is abduced, and the $j$-th body literal of $C_i$, for each abduced $use(i,j)$ atom. All other clauses and literals from $K_v$ are discarded. The bias towards hypotheses with fewer literals is realized by means of abducing a minimal set of $use/2$ atoms.

\noindent \begin{myexample}[\textbf{Example \ref{ex:xhail_functionality_1} continued}]
\label{ex:xhail_functionality_2}
$\Delta_2$ presented next to the ALP task of Phase 3 in Table \ref{table:xhail_functionality} is a minimal abductive explanation for this ALP task. $\mathit{use(1,0)}$ and $\mathit{use(2,0)}$ correspond to the head atoms of the two $K_v$ clauses, while $\mathit{use(1,3)}$ and $\mathit{use(2,2)}$ correspond respectively to their third and second body literal. The output hypothesis in Table \ref{table:xhail_functionality} is constructed by these literals, while all other literals and clauses from $K_v$ are discarded.   
\end{myexample}

To sum up, XHAIL provides an appropriate framework for learning event definitions in the form of Event Calculus programs. However, a serious obstacle that prevents XHAIL from being widely applicable as a machine learning system for event recognition is scalability. XHAIL scales poorly,  partly because of the increased computational complexity of adbuction, which lies at the core of its functionality, and partly because of the combinatorial complexity of learning whole theories, which may result in an intractable search space. In what follows, we use the XHAIL machinery to develop an incremental algorithm that scales to large volumes of sequential data, typical of event-based applications.

\section{ILED: Incremental Learning of Event Definitions}
\label{sec:incremental_learning}

We begin the presentation of our approach, which we call ILED (Incremental Learning of Event Definitions), by defining the incremental setting we assume and elaborating on the main challenges that stem from this setting. We then present the basic ideas that allow to address these challenges and proceed with a detailed description of the method.

\begin{definition}[Incremental Learning]
\label{def:incremental_learning}
We assume an ILP task \linebreak $\mathit{ILP(\mathsf{SDEC},\mathcal{E},M)}$, where $\mathcal{E}$ is a database of examples, called historical memory, storing examples presented over time. Initially $\mathcal{E} = \emptyset$. At time $n$ the learner is presented with a hypothesis $H_n$ such that $\mathsf{SDEC} \cup H_n \vDash \mathcal{E}$, in addition to a new set of examples $w_n$. The goal is to revise $H_n$ to a hypothesis $H_{n+1}$, so that $\mathsf{SDEC} \cup H_{n+1} \vDash \mathcal{E}\cup w_n$.
\end{definition}

A main challenge of adopting a full memory approach is to scale it up to a growing size of experience. This is in line with a key requirement of incremental learning where ``the incorporation of experience into memory during learning should be computationally efficient, that is, theory revision must be efficient in fitting new incoming observations'' \citep{langley_incremental_learning,avoid_order_effects}. In the stream processing literature, the number of passes over a stream of data is often used as a measure of the efficiency of algorithms \citep{li2004efficient,li2009mining}. In this spirit, the main contribution of ILED, in addition to scaling up XHAIL, is that it adopts a ``single-pass'' theory revision strategy, that is, a strategy that requires at most one pass over $\mathcal{E}$ in order to compute $H_{n+1}$ from $H_{n}$.

A single-pass revision strategy is far from trivial. For instance, the addition of a new clause $C$ in response to a set of new examples $w_n$ implies that $H_n$ must be checked throughout $\mathcal{E}$. In case $C$ covers some negative examples in $\mathcal{E}$ it should be specialized, which in turn may affect the initial coverage of $C$ in $w_n$. If the specialization results in the rejection of positive examples in $w_n$, extra clauses must be generated and added to $H_n$, in order to retrieve the lost positives, and these clauses should be again checked for correctness in $\mathcal{E}$. This process continues until a hypothesis $H_{n+1}$ is found, that accounts for all the examples in $\mathcal{E}\cup w_n$. In general, this requires several passes over the historical memory. 

Since experience may grow over time to an extent that is impossible to maintain in the working memory, we follow an external memory approach \citep{inthelex_database}. This implies that the learner does not have access to all past experience as a whole, but to independent sets of training data, in the form of \emph{sliding windows}. Sliding windows should be sufficiently large to capture the temporal dependencies between the data, as imposed by the \textsf{SDEC} axioms, which make the truth value of a fluent at time $T$ depend on what happens at $T{-}1$. We thus assume that sliding windows consist of at least two consecutive examples. For instance, the data in Table \ref{table:stream1} may be considered as part of two windows, or as part of a single window.

\scriptsize
\begin{algorithm}[t]
 \caption{$\mathtt{iled}(\mathsf{SDEC},M,H_n,w_n)$\newline
\textbf{Input:} \textit{The axioms of \textsf{SDEC}, mode declarations M, a hypothesis $H_n$ such that $\mathsf{SDEC}\cup H_n \vDash \mathcal{E}$} and an example window $w_n$.\newline
\textbf{Output:} \emph{A hypothesis $H_{n+1}$ such that $\mathsf{SDEC}\cup H_{n+1} \vDash \mathcal{E}\cup w_n$}}
 \label{alg:iled_overall_strategy} 
\begin{algorithmic}[1]

\If{$\mathsf{SDEC}\cup H_n \nvDash w_n$}
  \State \textbf{let} $K_v^{w_n}$ be a (variabilized) Kernel Set of $w_n$ \label{algline:iled-alg-line-ref-3}
  \State \textbf{let} $\mathit{\langle RetainedClauses,RefinedClauses,NewClauses \rangle} \leftarrow \mathsf{revise}(\mathsf{SDEC},H_n,K_{v}^{w_n},w_n)$ \label{algline:iled-alg-line-ref-4}
  \State \textbf{let} $\mathit{H' \leftarrow H_{keep}\cup RefinedClauses \cup NewClauses}$ \label{algline:iled-alg-line-ref-5}
  \If{$\mathit{NewClauses} \neq \emptyset$} \label{algline:iled-alg-line-ref-6}
    \ForAll{$w_i \in \mathcal{E}, \ 0\leq i \leq n-1$} \label{algline:iled-alg-line-ref-7}
      \If{$\mathsf{SDEC}\cup H' \nvDash w_i$}
        \State \textbf{let} $\mathit{\langle RetainedClauses,RefinedClauses,\emptyset \rangle} \leftarrow \mathsf{revise}(\mathsf{SDEC},H',\emptyset,w_i)$\label{algline:iled-alg-line-ref-9}
        \State \textbf{let} $\mathit{H' \leftarrow RetainedClauses\cup RefinedClauses}$         
      \EndIf     \label{algline:iled-alg-line-ref-8}
    \EndFor  
  \EndIf
  \State \textbf{let} $\mathit{H_{n+1} \leftarrow H'} $   
\Else
  \State \textbf{let} $H_{n+1} \leftarrow H_n$ \label{algline:iled-alg-line-ref-1} 
\EndIf     

\State \textbf{let} $\mathcal{E} \leftarrow \mathcal{E}\cup w_n$
\State \textbf{Return} $H_{n+1}$
\end{algorithmic}
\end{algorithm}
\normalsize

ILED's high-level strategy is presented in Algorithm \ref{alg:iled_overall_strategy}. At time $n$, ILED is presented with a hypothesis $H_n$ that accounts for the historical memory so far, and a new example window $w_n$. If the hypothesis at hand covers the new window then it is returned as is (line \ref{algline:iled-alg-line-ref-1}), otherwise ILED starts the process of revising $H_n$ (line \ref{algline:iled-alg-line-ref-4}). Revision operators that retract knowledge, such as the deletion of clauses or antecedents are excluded, due to the exponential cost of backtracking in the historical memory \citep{badea_2001}. The supported revision operators are thus:

\begin{itemize}
\item Addition of new clauses.
\item Refinement of existing clauses, i.e. replacement of an existing clause with one or more specializations of that clause. 
\end{itemize}

\noindent To treat incompleteness we add \myinitiatedAt \ clauses and refine \myterminatedAt \  clauses, while to treat inconsistency we add \myterminatedAt \ clauses and refine \myinitiatedAt \ clauses. 

\begin{figure}[t]
\centering
\includegraphics[width=1\textwidth]{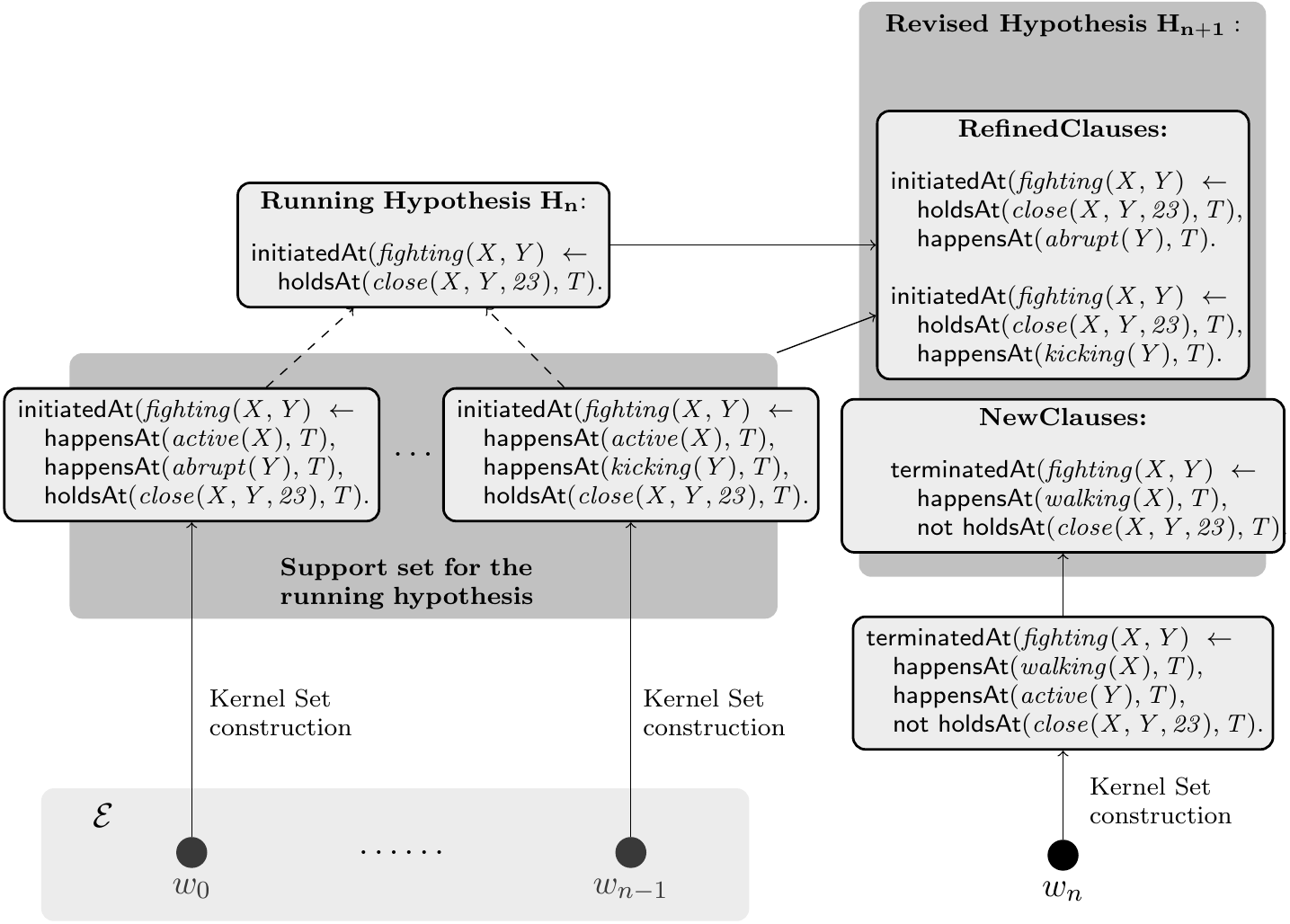}
\caption{Revision of a hypothesis $H_n$ in response to a new example window $w_n$. $\mathcal{E}$ represents the historical memory of examples.}
\label{fig:revision_schema}

\end{figure} 

Given a running hypothesis $H_n$ and a new window $w_n$, the goal of Algorithm \ref{alg:iled_overall_strategy} is to retain the \emph{preservable} clauses of $H_n$ intact, refine its \emph{revisable} clauses and, if necessary, generate a set of new clauses that account for new examples in the incoming window $w_n$. Definition \ref{def:preservable_revisable} provides a formal account for preservable and revisable clauses.

\begin{definition}[\textbf{Revisable and Preservable Parts of a Hypothesis}]
\label{def:preservable_revisable}
Let $H$ be a hypothesis, $C \in H$ a clause and $w$ an example window. We say that $C$ is revisable w.r.t $w$ if $\mathsf{SDEC\cup C}$ covers some negative examples or disproves some positive examples in $w$. Otherwise, we say that $C$ is preservable w.r.t. $w$.
\end{definition}

\noindent Revisions are implemented via the \textsf{revise} function (see line \ref{algline:iled-alg-line-ref-4} of Algorithm \ref{alg:iled_overall_strategy}). Figure \ref{fig:revision_schema} illustrates this function with a simple example. New clauses are generated by generalizing a Kernel Set of the incoming window, as shown in Figure \ref{fig:revision_schema}, where a \myterminatedAt/2 clause is generated from the new window $w_n$. Moreover, to facilitate refinement of existing clauses, each clause in the running hypothesis is associated with a memory of the examples it covers throughout $\mathcal{E}$, in the form of a ``bottom program'', which we call \emph{support set}. The support set is constructed gradually, as new example windows arrive. It serves as a refinement search space, as shown in Figure \ref{fig:revision_schema}, where the single clause in the running hypothesis $H_n$ is refined w.r.t. the incoming window $w_n$ into two specializations. Each such specialization results by adding to the initial clause one antecedent from the two support set clauses which are presented in Figure \ref{fig:revision_schema}. The revised hypothesis $H_{n+1}$ is constructed from the refined clauses and the new ones, along with the preserved clauses of $H_n$, if any (line \ref{algline:iled-alg-line-ref-5}, Algorithm \ref{alg:iled_overall_strategy}).



The historical memory is reconsulted only when new clauses are generated from the Kernel Set of the new window $w_n$ (see line \ref{algline:iled-alg-line-ref-6} of Algorithm \ref{alg:iled_overall_strategy}). The new clauses are checked on each example window in $\mathcal{E}$ and refined if necessary. At this step there is no need to generate new clauses, but only to ensure that the ones generated at the new window $w_n$ are consistent throughout $\mathcal{E}$. This is why in line \ref{algline:iled-alg-line-ref-9} of Algorithm \ref{alg:iled_overall_strategy}, $\mathit{NewClauses}$ and the Kernel Set $K$ are both empty in the result and the arguments of the \textsf{revise} function respectively. 

There are two key features of ILED that contribute towards its scalability: First, the re-processing of past experience is necessary only in the case where new clauses are generated and is redundant in the case where a revision consists of refinements of existing clauses. Second, as shown by the iteration of lines \ref{algline:iled-alg-line-ref-7}-\ref{algline:iled-alg-line-ref-8} of Algorithm \ref{alg:iled_overall_strategy}, re-processing of past experience requires a single pass over the historical memory, meaning that it suffices to ``re-see'' each past window exactly once to ensure that the output revised hypothesis $H_{n+1}$ is complete \& consistent w.r.t. the entire historical memory. These properties of ILED are due to the support set, which we present in detail.


\subsection{Support Set}
\label{sec:support_set}
The intuition behind the support set stems from the XHAIL methodology. Given a set of examples $E$, XHAIL learns a hypothesis by generalizing a Kernel Set $K$ of these examples. $E$ may be too large to process in one go and a possible solution is to partition $E$ in smaller example sets $E_1,\ldots,E_n$ and try to learn a hypothesis that accounts for the whole of $E$, by gradually revising an initial hypothesis $H_1$ acquired from $E_1$. In this process of progressive revisions, a compressive memory of ``small'' Kernel sets of $E_1,\ldots,E_n$ may be used as a  surrogate for the fact that one is not able to reason with the whole Kernel Set $K$. This is the role of the support set. 

By means of this memory, and as far as clause refinement is concerned, ILED is able to repair problems locally, i.e. in a single example window, without affecting coverage in the parts of the historical memory where the clause under refinement has been previously checked and is preservable. In more detail, given a hypothesis clause $C$ and a window $w$ where $C$ must be refined, and denoting by $\mathit{\mathcal{E}_{pr}(C)}$, the part of $\mathcal{E}$ where we know that $C$ is preservable, ILED refines $C$ so that its refinement covers all positive examples that $C$ covers in $\mathit{\mathcal{E}_{pr}(C)}$, making the task of checking $\mathit{\mathcal{E}_{pr}(C)}$ in response to the refinement redundant. 

In order to formally define the properties of the proposed memory structure, we use the notions of a \emph{depth-bound mode language} and \emph{most-specific clause}. Intuitively, given a set of mode declarations $M$ and a non-negative integer $i$, a clause $C$ is in the depth-bounded mode language $\mathcal{L}_i(M)$ if it is in the mode language $\mathcal{L}(M)$ and additionally, its length is bound by $i$. A clause $C$ in $\mathcal{L}_i(M)$ is most-specific if it does not $\theta$-subsume any other clause in $\mathcal{L}_i(M)$. These notions are formally defined in Definitions \ref{def:depth_mode_language} and \ref{def:most_specific_clause} in Appendix \ref{app:appendix_lp}. Definition \ref{def:additional_notation} provides some additional notation that we henceforth use.

\begin{definition}[\textbf{Notation}]
\label{def:additional_notation}
Let $\mathcal{E}$ be the historical memory, $M$ a set of mode declarations, $\mathcal{L}_i(M)$ the depth-bound mode language of $M$ for some non-negative integer $i$, $H\in \mathcal{L}_i(M)$ a running hypothesis and $C\in H$ a hypothesis clause. We use the following notation:
\begin{enumerate}[label=(\roman*)]
\item $cov_{\mathcal{E}}(C) = \{e \in \mathcal{E} \ | \ \mathsf{SDEC}\cup C \vDash e\}$, i.e. $cov_{\mathcal{E}}(C)$ denotes the coverage of clause $C$ in the historical memory. 
\item Given $E \subseteq \mathcal{E}$, $\mathcal{L}_i(M,E) = \{D \in \mathcal{L}_i(M) \ | \ \mathsf{SDEC} \cup D \vDash E\}$, i.e. $\mathcal{L}_i(M,E)$ denotes the fragment of the depth-bound mode language $\mathcal{L}_{i}(M)$ that covers a given set of examples $E$.  
\end{enumerate}
\end{definition}

\noindent Definition \ref{def:support_set} defines formally the properties of the support set.



\begin{definition}[\textbf{Support Set}]
\label{def:support_set}
Let $\mathcal{E}$, $M$ and $\mathcal{L}_i(M)$ be as in Definition \ref{def:additional_notation}, $H_n\in \mathcal{L}_i(M)$ be as in the Incremental Learning setting (Definition \ref{def:incremental_learning}) and let $C\in H$. The support set of $C$ is a program $\mathit{C.supp}$ with the following properties: 
\begin{enumerate}[label=(\roman*)]
\item $C \preceq D$ for each $\mathit{D \in C.supp}$.
\item Each $\mathit{D \in C.supp}$ is a most-specific clause of $\mathcal{L}_i(M,cov_{\mathcal{E}}(C))$.
\item $\mathit{cov_{\mathcal{E}}(C.supp)=cov_{\mathcal{E}}(C)}$. 
\end{enumerate}

\end{definition}

\noindent Properties (i) and (ii) of Definition \ref{def:support_set} imply that clause $C$ and its support set $\mathit{C.supp}$ define a space $\mathcal{S}$ of specializations of $C$, each of which is bound by a most-specific specialization, among those that cover the positive examples that $C$ covers, and up to a maximal clause length. In other words, for every $D \in \mathcal{S}$ there is a $\mathit{C_s \in C.supp}$ so that $C \preceq D \preceq C_s$ and $C_s$ covers at least one example from $\mathit{cov_{\mathcal{E}}(C)}$. Property (iii) of Definition \ref{def:support_set} ensures that space $\mathcal{S}$ contains refinements of clause $C$ that collectively preserve that coverage of $C$ in the historical memory. 
The purpose of $\mathit{C.supp}$ is thus to serve as a search space for refinements $R_{C}$ of clause $C$ for which $\mathit{C\preceq R_C \preceq C.supp}$ holds. In this way, clause $C$ may be refined w.r.t. a window $w_n$, avoiding the overhead of re-testing the refined program on $\mathcal{E}$. However, to ensure that the support set can indeed be used as a refinement search space, one must ensure that $\mathit{C.supp}$ will always contain such a refinement $R_C$, i.e. a preservable program w.r.t. a given window $w_n$, that may replace $C$ in case that latter is revisable w.r.t. $w_n$. Proposition \ref{prop:ss_can_be_used_as_refinement_search_space} shows that this is indeed the case.


\begin{proposition}
\label{prop:ss_can_be_used_as_refinement_search_space}
Let $H_n \in \mathcal{L}_i(M)$ be as in the Incremental Learning setting (Definition 4), i.e. $\mathsf{SDEC} \cup H_n \vDash \mathcal{E}$, and $w_n$ be an example window. Assume also that there exists a hypothesis $H_{n+1} \in \mathcal{L}_i(M)$, such that $\mathsf{SDEC} \cup H_{n+1} \vDash \mathcal{E} \cup w_n$, and that a clause $C\in H_n$ is revisable w.r.t. window $w_n$. Then $\mathit{C.supp}$ contains a refinement $R_C$ of $C$, which is preservable w.r.t. $w_n$. 
\end{proposition}
\begin{myproof}
Assume, towards contradiction, that each each refinement $R_C$ of $C$, contained in $\mathit{C.supp}$ is revisable w.r.t. $w_n$. It then follows that $\mathit{C.supp}$ itself is  revisable w.r.t. $w_n$, i.e. it either covers some negative examples, or it disproves some positive examples in $w_n$. Let $e_{1}\in w_n$ be such an example that $\mathit{C.supp}$ fails to satisfy, and assume for simplicity that a single clause $C_s \in \mathit{C.supp}$ is responsible for that. By definition, $C_s$ covers at least one positive example $e_{2}$ from $\mathcal{E}$ and furthermore, it is a most-specific clause, within $\mathcal{L}_{i}(M)$, with that property. It then follows that $e_{1}$ and $e_{2}$ cannot both be accounted for, under the given language bias $\mathcal{L}_i(M)$, i.e. there exists no hypothesis $H_{n+1} \in \mathcal{L}_i(M)$ such that $\mathsf{SDEC} \cup H_{n+1} \vDash \mathcal{E} \cup w_n$, which contradicts our assumption. Hence $\mathit{C.supp}$ is preservable w.r.t. $w_n$ and it thus contains a refinement $R_{C}$ of $C$, which is preservable w.r.t. $w_n$.    
\end{myproof}


The construction of the support set, presented in Algorithm \ref{alg:support_construction}, is a process that starts when $C$ is added in the running hypothesis and continues as long as new example windows arrive. While this happens, clause $C$ may be refined or retained, and its support set is updated accordingly. The details of Algorithm \ref{alg:support_construction} are presented in Example \ref{ex:supported_refinement_example}, which also demonstrates how ILED processes incoming examples and revises hypotheses. 


\scriptsize
\begin{algorithm}[t]
 \caption{Support set construction and maintenance}
 \label{alg:support_construction} 
\begin{algorithmic}[1]
\State \begin{varwidth}[t]{\linewidth} \textbf{let} $w_n\notin \mathcal{E}$ be an example window, $H_n$ a current hypothesis and \par $\mathit{H'_n=NewClauses\cup RefinedClauses \cup RetainedClauses}$ a revision of $H_n$, generated in $w_n$, where $\mathit{NewClauses, RefinedClauses}$ and $\mathit{RetainedClauses}$ are as described in Algorithm \ref{alg:iled_overall_strategy}. \end{varwidth}
\ForAll{$C\in H'_{n}$}
  \If{$\mathit{C\in NewClauses}$} \label{algline:ss-alg-ref-3}
    \State \begin{varwidth}[t]{\linewidth} $\mathit{C.supp \leftarrow \{D \in K \ | \ C \preceq D\}}$, where $K$ is the \textbf{variabilized Kernel Set} of $w_n$ \par from which $\mathit{NewClauses}$ is generated. \end{varwidth}
  \ElsIf{$\mathit{C \in RefinedClauses}$} \label{algline:ss-alg-ref-4}
    \State \begin{varwidth}[t]{\linewidth} $\mathit{C.supp \leftarrow \{D \in C_{parent}.supp \ | \ C \preceq D\}}$, where $\mathit{C_{parent}}$ is the ``ancestor'' \par clause of $C$, i.e. the clause from which $C$ results by specialization.
    \end{varwidth}
  \Else \label{algline:ss-alg-ref-5}
    \State \begin{varwidth}[t]{\linewidth} \textbf{let} $e_{C}^{w_n}$ be the \emph{true positives} that $C$ covers in $w_n$, if $C$ is an \textsf{initiatedAt} clause, or \par
\hskip\algorithmicindent  the \emph{true negatives} that $C$ covers, if it is a  \textsf{terminatedAt} clause. 
\end{varwidth}  
\If{$\mathsf{SDEC}\cup \mathit{C.supp} \nvDash e_{C}^{w_n}$}
\State \textbf{let} $K$ be a \textbf{variabilized Kernel Set} of $w_n$.
\State $\mathit{C.supp} \leftarrow \mathit{C.supp} \cup K'$, where $K'\subseteq K$, such that $\mathsf{SDEC}\cup K' \vDash e_{C}^{w_n}$ \label{algline:ss-alg-ref-6} 
\EndIf 
  \EndIf    
\EndFor

\end{algorithmic}
\end{algorithm}
\normalsize

\begin{table}[t]
\scriptsize
\begin{center}
\setlength\tabcolsep{3pt}
\begin{tabular}{ll}
\noalign{\smallskip}
\hline
\noalign{\smallskip}
\multicolumn{1}{l}{\textbf{Window $w_1$}}   \\
\hline\noalign{\smallskip}

\textbf{Narrative}  & \textbf{Annotation} \\
 \  \ \myhappens($\mathit{active(id_1),10}$). & \  \  \mynbf \ \myholdsAt($\mathit{fighting(id_1,id_2),10}$).\\
 \  \ \myhappens($\mathit{abrupt(id_2),10}$). & \  \ \myholdsAt($\mathit{fighting(id_1,id_2),11}$). \\
 \  \ \myholdsAt($\mathit{close(id_1,id_2,23),10}$). &\\
~ & ~ \\

\textbf{Kernel Set}   & \textbf{Variabilized Kernel Set} \\
 \  \ $\myinitiatedAt(fighting(id_1,id_2),10) \leftarrow$   &   \  \ $K_1 = \myinitiatedAt(fighting(X,Y),T) \leftarrow$ \\
$\qquad \qquad \myhappens(active(id_1),10)$, &   \  \      $\qquad \qquad \myhappens(active(X),T)$,\\
$\qquad \qquad \myhappens(abrupt(id_2),10)$ &   \  \      $\qquad \qquad \myhappens(abrupt(Y),T)$,\\
$\qquad \qquad \myholdsAt(close(id_1,id_2,23),10)$ &   \   \      $\qquad \qquad \myholdsAt(close(X,Y,23),T)$.\\  

~ & ~ \\

\textbf{Running Hypothesis}   & \textbf{Support Set} \\
 \  \ $C = \myinitiatedAt(fighting(X,Y),T) \leftarrow$   &   \  \ $\mathit{C.supp = \{K_1\}}$ \\
$\qquad \qquad \myhappens(active(X),T).$ &   \  \     \\

\hline\noalign{\smallskip}
\multicolumn{1}{l}{\textbf{Window $w_2$}}   \\
\hline\noalign{\smallskip}

\textbf{Narrative}  & \textbf{Annotation} \\
 \  \ \myhappens($\mathit{active(id_1),20}$). & \  \  \mynbf \ \myholdsAt($\mathit{fighting(id_1,id_2),20}$).\\
 \  \ \myhappens($\mathit{kicking(id_2),20}$). & \  \ \myholdsAt($\mathit{fighting(id_1,id_2),21}$). \\
 \  \ \myholdsAt($\mathit{close(id_1,id_2,23),20}$). &\\

~ & ~ \\

\textbf{Kernel Set}   & \textbf{Variabilized Kernel Set} \\
 \  \ $\myinitiatedAt(fighting(id_1,id_2),20) \leftarrow$   &   \  \ $K_2 = \myinitiatedAt(fighting(X,Y),T) \leftarrow$ \\
$\qquad \qquad \myhappens(active(id_1),20)$, &   \  \      $\qquad \qquad \myhappens(active(X),T)$,\\
$\qquad \qquad \myhappens(kicking(id_2),20)$ &   \  \      $\qquad \qquad \myhappens(kicking(Y),T)$,\\
$\qquad \qquad \myholdsAt(close(id_1,id_2,23),20)$ &   \   \      $\qquad \qquad \myholdsAt(close(X,Y,23),T)$.\\  

~ & ~ \\

\textbf{Running Hypothesis}   & \textbf{Support Set} \\
 \  \ Remains unchanged   &   \  \ $\mathit{C.supp = \{K_1,K_2\}}$ \\

\hline\noalign{\smallskip}
\multicolumn{1}{l}{\textbf{Window $w_3$}}   \\
\hline\noalign{\smallskip}

\textbf{Narrative}  & \textbf{Annotation} \\
 \  \ \myhappens($\mathit{active(id_1),30}$). & \  \  \mynbf \ \myholdsAt($\mathit{fighting(id_1,id_2),30}$).\\
 \  \ \myhappens($\mathit{walking(id_2),30}$). & \  \ \mynbf \ \myholdsAt($\mathit{fighting(id_1,id_2),31}$).\\
 \  \ \mynbf \ \myholdsAt($\mathit{close(id_1,id_2,23),30}$). &\\

~ & ~ \\

\textbf{Revised Hypothesis}   & \textbf{Support Set} \\
$C_1 = \myinitiatedAt(fighting(X,Y),T) \leftarrow$ & $\mathit{C_1.supp = \{K_1,K_2\}}$\\
$\qquad \qquad \myhappens(active(X),T),$ & \\
$\qquad \qquad \myholdsAt(close(X,Y,23),T).$      &  \\

\hline
\end{tabular}
\caption{\small Knowledge for Example \ref{ex:supported_refinement_example}}\label{table:support_example}
\end{center}
\end{table}
\normalsize

\noindent \begin{myexample}
\label{ex:supported_refinement_example}
Consider the annotated examples and running hypothesis related to the \emph{fighting} high-level event from the activity recognition application shown in Table \ref{table:support_example}. We assume that ILED starts with an empty hypothesis and an empty historical memory, and that $w_1$ is the first input example window. The currently empty hypothesis does not cover the provided examples, since in $w_1$ \emph{fighting} between persons $id_1$ and $id_2$ is initiated at time 10 and thus holds at time 11. Hence ILED starts the process of generating an initial hypothesis. In the case of an empty hypothesis,  ILED reduces to XHAIL and operates on a Kernel Set of $w_1$ only, trying to induce a minimal program that accounts for the examples in $w_1$. The variabilized Kernel Set in this case will be the single-clause program $K_1$ presented in Table \ref{table:support_example} generated from the corresponding ground clause. Generalizing this Kernel Set yields a minimal hypothesis that covers $w_1$. One such hypothesis is clause $C$ shown in Table \ref{table:support_example}. ILED stores $w_1$ in $\mathcal{E}$ and initializes the support set of the newly generated clause $C$ as in line \ref{algline:ss-alg-ref-3} of Algorithm \ref{alg:support_construction}, by selecting from $K_1$ the clauses that are $\theta$-subsumed by $C$, in this case, $K_1$'s single clause. 

Window $w_2$ arrives next. In $w_2$, \emph{fighting} is initiated at time 20 and thus holds at time 21. The running hypothesis correctly accounts for that and thus no revision is required. However, $\mathit{C.supp}$ does not cover $w_2$ and unless proper actions are taken, property (iii) of Definition \ref{def:support_set} will not hold once $w_2$ is stored in $\mathcal{E}$. ILED thus generates a new Kernel Set $K_2$ from window $w_2$, as presented in Table \ref{table:support_example}, and updates $\mathit{C.supp}$ as shown in lines \ref{algline:ss-alg-ref-5}-\ref{algline:ss-alg-ref-6} of Algorithm \ref{alg:support_construction}. Since $C$ $\theta$-subsumes $K_2$, the latter is added to $\mathit{C.supp}$, which now becomes $\mathit{C.supp} = \{K_1,K_2\}$. Now $\mathit{cov_{\mathcal{E}}(C.supp) = cov_{\mathcal{E}}(C)}$, hence in effect, $\mathit{C.supp}$ is a summarization of the coverage of clause $C$ in the historical memory.

Window $w_3$ arrives next, which has no positive examples for the initiation of \emph{fighting}. The running hypothesis is revisable in window $w_3$, since clause $C$ covers a negative example at time 31, by means of initiating the fluent $\mathit{fighting(id_1,id_2)}$ at time 30. To address the issue, ILED searches $\mathit{C.supp}$, which now serves as a refinement search space, to find a refinement $R_{C}$ that rejects the negative example, and moreover $\mathit{R_C \preceq C.supp}$. Several choices exist for that. For instance, the following program

\scriptsize
\noindent\begin{tabularx}{\textwidth}{@{}XXX@{}}
  \begin{equation*}
    \label{eq:ec_axioms_1}
    \begin{array}{lr}
    \myinitiatedAt(fighting(X,Y),T) \leftarrow \\
     \qquad \myhappens(active(X),T),\\
     \qquad \myhappens(abrupt(Y),T).\\    
    \end{array}
  \end{equation*} &
    \begin{equation*}
    \label{eq:ec_axioms_1}
    \begin{array}{lr}
     \myinitiatedAt(fighting(X,Y),T) \leftarrow \\
     \qquad \myhappens(active(X),T),\\
     \qquad \myhappens(kicking(Y),T).\\
    \end{array}
  \end{equation*}
\end{tabularx}  
\normalsize

\noindent is such a refinement $R_C$, since it does not cover the negative example in $w_3$ and subsumes $\mathit{C.supp}$. ILED however is biased towards minimal theories, in terms of the overall number of literals and would prefer the more compressed refinement $C_1$, shown in Table \ref{table:support_example}, which also rejects the negative example in $w_3$ and subsumes $\mathit{C.supp}$. Clause $C_1$ replaces the initial clause $C$ in the running hypothesis. The hypothesis now becomes complete and consistent throughout $\mathcal{E}$. Note that the hypothesis was refined by local reasoning only, i.e. reasoning within window $w_3$ and the support set, avoiding costly look-back in the historical memory. The support set of the new clause $C_1$ is initialized (line \ref{algline:ss-alg-ref-4} of Algorithm \ref{alg:support_construction}), by selecting the subset of the support set of its parent clause that is $\theta$-subsumed by $C_1$. In this case $\mathit{C_1 \preceq C.supp = \{K_1,K_2\}}$, hence $\mathit{C_1.supp = C.supp}$.

\normalsize
\end{myexample}


\noindent As shown in Example \ref{ex:supported_refinement_example}, the support set of a clause $C$ is a compressed enumeration of the examples that $C$ covers throughout the historical memory. It is compressed because it is expected to encode many examples with a single variabilized clause. In contrast, a ground version of the support set would be a plain enumeration of examples, since in the general case, it would require one ground clause per example. The main advantage of the ``lifted'' character of the support set over a plain enumeration of the examples is that it requires much less memory to encode the necessary information, an important feature in large-scale (temporal) applications. Moreover, given that training examples are typically characterized by heavy repetition, abstracting away redundant parts of the search space results in a memory structure that is expected to grow in size slowly, allowing for fast search that scales to large amount of historical data. 

\subsection{Implementing Revisions}
\label{sec:operator_implementation}

\scriptsize
\begin{algorithm}[t]
\caption{$\mathsf{revise(SDEC},H_n,w_n,K_{v}^{w_n})$\newline
\textbf{Input:} \textit{The axioms of $\mathsf{SDEC}$, a running hypothesis $H_n$ an example window $w_n$ and a variabilized Kernel Set $K_{v}^{w_n}$ of $w_n$}.\newline
\textbf{Output:} \textit{A revised hypothesis} $H_{n}'$
}
 \label{alg:revise} 
\begin{algorithmic}[1]

\State \textbf{let} $\mathit{U(K_{v}^{w_n},H_n) \leftarrow \mathsf{GeneralizationTransformation}(K_{v}^{w_n})\cup \mathsf{RefinementTransformation}(H_n)}$  \label{line:revise_syntactic_trasnf}

\State \textbf{let} $\Phi$ be the abductive task $\Phi=\mathit{ALP(\mathsf{SDEC}\cup U(K_{v}^{w_n},H_n),\{use/2,use/3\},w_n)}$ \label{line:revise_first_abduction}

\If{$\Phi$ has a solution}
\State \textbf{let} $\Delta$ be a minimal solution of $\Phi$

\State $\begin{array}{lr}
    \text{\textbf{let} }\mathit{NewClauses = \{\alpha_i \leftarrow \delta_{i}^{1} \wedge \ldots \wedge \delta_{i}^{n} \ | }\\
    \qquad \qquad \qquad \qquad \quad  \alpha_i \text{ is the head of the } i{-}\text{th clause } C_i \in K_{v}^{w_n} \\
    \qquad \qquad \qquad \qquad \quad  \text{and } \delta_{i}^{j} \text{ is the } j{-}\text{th body literal of } C_i \\
     \mathit{\qquad \qquad \qquad \quad \qquad  \text{and } use(i,0) \in \Delta \text{ and } use(i,j) \in \Delta, 1\leq j \leq n \ \}}\\
     \end{array}$\label{line:new_clauses}

\State $\begin{array}{lr}
    \text{\textbf{let} }\mathit{RefinedClauses = \{ \ head(C_i) \leftarrow body(C_i) \wedge \delta_{i}^{j,k_1} \wedge \ldots \wedge \delta_{i}^{j,k_m} \ | }\\
     \mathit{\qquad \qquad \qquad \qquad \qquad   \ C_i \in H_n \text{ and } use(i,j,k_{l}) \in \Delta, 1\leq l \leq m, 1\leq j \leq |C_i.supp| \ \}}\\
     \end{array}$\label{line:refined_clauses}

\State \textbf{let} $\mathit{RetainedClauses = \{C_i \in H_n \ | \ use(i,j,k)\notin \Delta \text{ for any } j,k \}}$\label{line:retained_clauses}

\State \textbf{let} $\mathit{RefinedClauses = \mathsf{ReduceRefined}(NewClauses,RefinedClauses,RetainedClauses)}$\label{line:reduce_refined}

\Else
\State \textbf{Return} \texttt{No Solution}
\EndIf
\State \textbf{Return} $\mathit{\langle RetainedClauses,RefinedClauses,NewClauses \rangle}$

\end{algorithmic}
\end{algorithm}
\normalsize

Algorithm \ref{alg:revise} presents the details of the \textsf{revise} function from Algorithm \ref{alg:iled_overall_strategy}. The input consists of \textsf{SDEC} as background knowledge, a running hypothesis $H_n$, an example window $w_n$ and a variabilized Kernel Set $K_{v}^{w_n}$ of $w_n$. The clauses of $K_{v}^{w_n}$ and $H_n$ are subject to the \textsf{GeneralizationTansformation} and the \textsf{RefinementTransformation} respectively, presented in Table \ref{table:syntactic_transforms}. The former is the transformation discussed in Section \ref{sec:xhail}, that turns the Kernel Set into a defeasible program, allowing to construct new clauses from the Kernel Set select, in order to cover the examples. The \textsf{RefinementTransformation} aims at the refinement of the clauses of $H_n$ using their support sets. It involves two fresh predicates, $\mathit{exception/3}$ and $use/3$. For each clause $D_i \in H_n$ and for each of its support set clauses $\Gamma_{i}^{j} \in D_i.supp$, one new clause $\mathit{head(D_i) \leftarrow body(D_i) \wedge \mynbf \ exception(i,j,v(head(D_i)))}$ is generated, where $\mathit{v(head(D_i))}$ is a term that contains the variables of $\mathit{head(C_i)}$. Then an additional clause $\mathit{exception(i,j,v(head(D_i))) \leftarrow use(i,j,k)\wedge \mynbf \ \delta_{i}^{j,k}}$ is generated, for each body literal $\delta_{i}^{j,k} \in \Gamma_{i}^{j}$. 

The syntactically transformed clauses are put together in a program \linebreak $U(K_{v}^{w_n},H_n)$ (line \ref{line:revise_syntactic_trasnf} of Algorithm \ref{alg:revise}), which is used as a background theory along with \textsf{SDEC}. A minimal set of $use/2$ and $use/3$ atoms is abduced as a solution to the abductive task $\Phi$ in line \ref{line:revise_first_abduction} of Algorithm \ref{alg:revise}. Abduced $\mathit{use/2}$ atoms are used to construct a set of $\mathit{NewClauses}$, as discussed in Section \ref{sec:xhail} (line \ref{line:new_clauses} of Algorithm \ref{alg:revise}). These new clauses account for some of the examples in $w_n$, which cannot be covered by existing clauses in $H_n$. The abduced $\mathit{use/3}$ atoms indicate clauses of $H_n$ that must be refined. From these atoms, a refinement $R_{D_i}$ is generated for each incorrect clause $D_i \in H_n$, such that $D_i \preceq R_{D_i} \preceq D_i.supp$ (line \ref{line:refined_clauses} of Algorithm \ref{alg:revise}). Clauses that lack a corresponding $use/3$ atom in the abductive solution are retained (line \ref{line:retained_clauses} of Algorithm \ref{alg:revise}).

\begin{table}[t]
\scriptsize
\begin{center}
\renewcommand{\arraystretch}{0.9}
\setlength{\tabcolsep}{5pt}
\begin{tabular}{ll}
\noalign{\smallskip}
\hline
\noalign{\smallskip}
\multicolumn{1}{l}{\textbf{\textsf{GeneralizationTransformation}}} & \multicolumn{1}{l}{\textbf{\textsf{RefinementTransformation}}}  \\
~ & ~\\
\textbf{Input:} A variabilized Kernel set $K_{v}$  & \textbf{Input:} A running hypothesis $H_n$  \\ 
\hline\noalign{\smallskip}
~ & ~ \\
\textbf{For each} clause $\mathit{D_i = \alpha_{i} \leftarrow \delta_{i}^{1},\ldots,\delta_{i}^{n}\in F_v}$: & \textbf{For each} clause $\mathit{D_i\in H_n}$:\\
~ Add an extra atom $\mathit{use(i,0)}$ to the body of $D_i$  & ~ \textbf{For each} clause $\mathit{\Gamma_{i}^{j} \in D_i.supp}$     \\
~ and replace each body literal $\delta_{i}^{j}$ with a new & ~ ~ Generate one clause  \\
~ atom  of the form $\mathit{try(i,j,v(\delta_{i}^{j}))}$, where $\mathit{v(\delta_{i}^{j})}$  & ~ ~ $\mathit{\alpha_i \leftarrow body(D_i) \wedge \mynbf \ exception(i,j,v(\alpha_{i}))}$  \\
~ contains the variables that appear in $\delta_{i}^{j}$. & ~ ~ where $\alpha_i$ is the head of $D_i$ and $\mathit{v(\alpha_i)}$  \\
~ Generate two new clauses of the form & ~ ~ contains its variables. Generate one clause \\
~ $\mathit{try(i,j,v(\delta_{i}^{j})) \leftarrow use(i,j),\delta_{i}^{j}}$  and & ~ ~ $\mathit{exception(i,j,v(a_i)) \leftarrow use(i,j,k),\mynbf \ \delta_{i}^{j,k}}$ \\
~ $\mathit{try(i,j,v(\delta_{i}^{j})) \leftarrow \mynbf \ use(i,j)}$ for each $\delta_{i}^{j}$.& ~ ~ for each body literal $\delta_{i}^{j,k}$ of $\Gamma_{i}^{j}$.\\
~ & ~ \\
\hline
\end{tabular}
\caption{\small Syntactic transformations performed by ILED.}\label{table:syntactic_transforms}
\end{center}
\end{table}
\normalsize

The intuition behind refinement generation is as follows: Assume that clause $D_i\in H_n$ covers negative examples or disproves positive examples in window $w_n$. To prevent that, the negation of the exception atom that is added to the body of $D_i$ during the \textsf{RefinementTransformation}, must fail to be satisfied, hence the exception atom itself must be satisfied. This can be achieved in several ways by means of the extra clauses generated by the \textsf{RefinementTransformation}. These clauses provide definitions for the exception atom, namely one for each body literal in each clause of $D_i.supp$. From these rules one can satisfy the exception atom by satisfying the complement of the corresponding support set literal and abducing the accompanying $\mathit{use/3}$ atom. In this way, each incorrect clause $D_i\in H_n$ and each $\mathit{\Gamma_{i}^{j} \in D_i.supp}$ correspond to a set of abduced $use/3$ atoms of the form $use(i,j,k_1),\ldots,use(i,j,k_n)$. These atoms indicate that a specialization of $D_i$ may be generated by adding to the body of $D_i$ the literals $\delta_{i}^{j,k_1},\ldots,\delta_{i}^{j,k_n}$ from $\Gamma_{i}^{j}$. Then a refinement $R_{D_i}$ such that $\mathit{D_i \preceq R_{D_i} \preceq D_i.supp}$ may be generated by selecting one specialization of clause $D_i$ from each support set clause in $\mathit{D_i.supp}$.

\begin{table}[t]
\scriptsize
\begin{center}
\renewcommand{\arraystretch}{0.9}
\setlength{\tabcolsep}{2pt}
\begin{tabular}{ll}
\noalign{\smallskip}
\hline
\noalign{\smallskip}
\multicolumn{1}{l}{\textbf{Input}} & \multicolumn{1}{l}{}  \\
\hline\noalign{\smallskip}
~ & ~ \\
\textbf{Narrative} & \textbf{Annotation}  \\
~ & ~ \\
$\mathit{\myhappens(abrupt(id_1), \ 1).}$ & $\mynbf \ \myholdsAt(fighting(id_1,id_2),1).$\\
$\mathit{\myhappens(inactive(id_2), \ 1).}$ & $\mynbf \ \myholdsAt(fighting(id_3,id_4),1).$  \\
$\mathit{\myholdsAt(close(id_1,id_2,23), \ 1).}$ & $\mynbf \ \myholdsAt(fighting(id_1,id_2),2).$ \\
$\mathit{\myhappens(abrupt(id_3), \ 2).}$ & $\mynbf \ \myholdsAt(fighting(id_3,id_4),2).$ \\
$\mathit{\myhappens(abrupt(id_4), \ 2).}$ &  $\mynbf \ \myholdsAt(fighting(id_1,id_2),3).$\\
$\mathit{\mynbf \ \myholdsAt(close(id_3,id_4,23), \ 2).}$ & $\mynbf \ \myholdsAt(fighting(id_3,id_4),3).$\\
~ & ~ \\
\textbf{Running hypothesis} & \textbf{Support set}  \\
~ & ~ \\
 $\mathit{C = \myinitiatedAt(fighting(X,Y), \ T)}\  \leftarrow$ & $\mathit{C_s^1 = \myinitiatedAt(fighting(X,Y), \ T)}\  \leftarrow$ \\
 $\qquad \quad \mathit{\myhappens(abrupt(X), \ T).} $ & $\qquad \quad \mathit{\myhappens(abrupt(X), \ T),} $ \\
 ~ & $\qquad \quad \mathit{\myhappens(abrupt(Y), \ T),} $\\
 ~ & $\qquad \quad \mathit{\myholdsAt(close(X,Y,23), \ T)}.$\\
 ~ & ~ \\
~ & $\mathit{C_s^2 = \myinitiatedAt(fighting(X,Y), \ T)}\  \leftarrow$ \\
~ & $\qquad \quad \mathit{\myhappens(abrupt(X), \ T),} $ \\
~ & $\qquad \quad \mathit{\myhappens(active(Y), \ T),} $ \\
~ & $\qquad \quad \mathit{\myholdsAt(close(X,Y,23), \ T)}.$\\
~ & ~ \\
\hline\noalign{\smallskip}
\textbf{Refinement transformation:} & ~ \\
\hline\noalign{\smallskip}
~ & ~ \\
\textbf{From} $C_{s}^1:$ & \textbf{From} $C_{s}^2:$ \\
~ & ~ \\
$\myinitiatedAt(fighting(X,Y), \ T) \leftarrow$ & $\myinitiatedAt(fighting(X,Y), \ T) \leftarrow$ \\
$\qquad \myhappens(abrupt(X), \ T),$ & $\qquad \myhappens(abrupt(X), \ T),$ \\
$\qquad \mynbf \ exception(1,1,vars(X,Y,T)).$ & $\qquad \mynbf \ exception(1,2,vars(X,Y,T)).$\\
$exception(1,1,vars(X,Y,T)) \leftarrow$ & $exception(1,2,vars(X,Y,T)) \leftarrow$ \\
$\qquad use(1,1,2),\mynbf \ \myhappens(abrupt(Y),T).$ & $\qquad use(1,2,2),\mynbf \ \myhappens(active(Y),T).$\\
$exception(1,1,vars(X,Y,T)) \leftarrow$ & $exception(1,2,vars(X,Y,T)) \leftarrow$ \\
$\qquad use(1,1,3),\mynbf \ \myholdsAt(close(X,Y,23),T).$ & $\qquad use(1,2,3), \mynbf \ \myholdsAt(close(X,Y,23),T).$ \\
~ & ~ \\

\hline\noalign{\smallskip}
\multicolumn{1}{l}{\textbf{Minimal abductive solution}} & \multicolumn{1}{l}{\textbf{Generated refinements}}  
\\ 
\hline\noalign{\smallskip}
 ~ & ~ \\
$\mathit{\Delta = \{use(1,1,2),use(1,1,3),use(1,2,2)\}}$ & $\mathit{\myinitiatedAt(fighting(X,Y), \ T)\  \leftarrow}$ \\
 ~ & $\quad  \mathit{\myhappens(abrupt(X), \ T),} $ \\
 ~ & $\quad  \mathit{\myhappens(abrupt(Y), \ T),} $\\
 ~ & $\quad  \mathit{\myholdsAt(close(X,Y,23), \ T)}.$\\
 ~ & ~ \\
~ & $\mathit{\myinitiatedAt(fighting(X,Y), \ T)}\  \leftarrow$ \\
~ & $\quad  \mathit{\myhappens(abrupt(X), \ T),} $ \\
~ & $\quad  \mathit{\myhappens(active(Y), \ T).} $ \\
 ~ & ~ \\
\hline
\end{tabular}
\caption{\small Clause refinement by ILED. }\label{table:refinement_example}
\end{center}
\end{table}
\normalsize

\noindent \begin{myexample}
\label{example:incons}
Table \ref{table:refinement_example} presents the process of ILED's refinement. The annotation lacks positive examples and the running hypothesis consists of a single clause $C$, with a support set of two clauses. Clause $C$ is inconsistent since it entails two negative examples, namely $\mathit{\myholdsAt(fighting(id_1,id_2), \ 2)}$ and \linebreak $\mathit{\myholdsAt(fighting(id_3,id_4), \ 3)}$. The program that results by applying the \textsf{RefinementTransformation} to the support set of clause $C$ is presented in Table \ref{table:refinement_example}, along with a minimal abductive explanation of the examples, in terms of $use/3$ atoms. Atoms $use(1,1,2)$ and $use(1,1,3)$ correspond respectively to the second and third body literals of the first support set clause, which are added to the body of clause $C$, resulting in the first specialization presented in Table \ref{table:refinement_example}. The third abduced atom $use(1,2,2)$ corresponds to the second body literal of the second support set clause, which results in the second specialization in Table \ref{table:refinement_example}. Together, these specializations form a refinement of clause $C$ that subsumes $\mathit{C.supp}$.

\end{myexample}

\noindent Minimal abductive solutions imply that the running hypothesis is minimally revised. Revisions are minimal w.r.t. the length of the clauses in the revised hypothesis, but are not minimal w.r.t. the number of clauses, since the refinement strategy described above may result in refinements that include redundant clauses: Selecting randomly one specialization from each support set clause to generate a refinement of a clause is sub-optimal, since there may exist other refinements with fewer clauses that also subsume the whole support set, as Example \ref{ex:supported_refinement_example} demonstrates. To avoid unnecessary increase at the hypothesis size, the generation of refinements is followed by a ``reduction'' step (line \ref{line:reduce_refined} of Algorithm \ref{alg:revise}). The \textsf{ReduceRefined} function works as follows. For each refined clause $C$, it first generates all possible refinements from $\mathit{C.supp}$. This can be realized with the abductive refinement technique described above. The only difference is that the abductive solver is instructed to find all abductive explanations in terms of $use/3$ atoms, instead of one. Once all refinements are generated, \textsf{ReduceRefined} searches the revised hypothesis, augmented with all refinements of clause $C$, to find a reduced set of refinements of $C$ that subsume $\mathit{C.supp}$. 

\subsection{Soundness and Single-Pass Theory Revision}
\label{sec:iled_correctness}

In this section we prove the correctness of ILED (Algorithm \ref{alg:iled_overall_strategy}) and show that it requires at most one pass over the historical memory to revise an input hypothesis.

\begin{proposition}[\textbf{Soundness and Single-pass Theory Revision}]
Assume the incremental learning setting described in Definition \ref{def:incremental_learning}. ILED (Algorithm \ref{alg:iled_overall_strategy}) requires at most one pass over $\mathcal{E}$ to compute $H_{n+1}$ from $H_n$.
\end{proposition}
\begin{myproof}
For simplicity and without loss of generality, we assume that when a new example window $w_n$ arrives, ILED revises $H_n$ by (a) refining an single clause $C\in H_n$ or (b) adding a new clause $C'$. 

In case (a), clause $C$ is replaced by a refinement $R_C$ such that $\mathit{C\preceq R_C \preceq C.supp}$. By property (iii) of the support set definition (Definition \ref{def:support_set}), $R_C$ covers all positive examples that $C$ covers in $\mathcal{E}$, hence for the hypothesis $H_{n+1} = (H_n \smallsetminus C) \cup R_{C}$ it holds that $\mathsf{SDEC} \cup H_{n+1} \vDash \mathcal{E}$ and furthermore $\mathsf{SDEC} \cup H_{n+1} \vDash w_n$. Hence $\mathsf{SDEC} \cup H_{n+1} \vDash \mathcal{E}\cup w_n$, from which soundness for $H_{n+1}$ follows. In this case $H_{n+1}$ is constructed from $H_n$ in a single step, i.e. by reasoning within $w_n$ without re-seeing other windows from $\mathcal{E}$.  

In case (b), $H_n$ is revised w.r.t. $w_n$ to a hypothesis $H_{n}' = H_{n} \cup C'$, where $C'$ is a new clause that results from the generalization of a Kernel Set of $w_n$. In response to the new clause addition, each window in $\mathcal{E}$ must be checked and $C'$ must be refined if necessary, as shown in line \ref{algline:iled-alg-line-ref-6} of Algorithm \ref{alg:iled_overall_strategy}. Let $\mathit{\mathcal{E}_{tested}}$ denote the fragment of $\mathcal{E}$ that has been tested at each point in time. Initially, i.e. once $C'$ is generated from $w_n$, it holds that $\mathit{\mathcal{E}_{tested}} = w_n$. At each window that is tested, clause $C'$ may (i) remain intact, (ii) be refined, or (iii) one of its refinements may be further refined. Assume that $w_k, \ k < n$ is the first window where the new clause $C'$ must be refined. At this point, $\mathit{\mathcal{E}_{tested} = \{ w_i \in \mathcal{E} \ | \ k < i \leq n\}}$, and it holds that $C'$ is preservable in $\mathit{\mathcal{E}_{tested}}$, since $C'$ has not yet been refined. In $w_k$, clause $C'$ is replaced by a refinement $R_{C'}$ such that $\mathit{C' \preceq R_{C'} \preceq C'.supp}$. $R_{C'}$ is preservable in $\mathit{\mathcal{E}_{tested}}$, since it is a refinement of a preservable clause, and furthermore, it covers all positive examples that $C'$ covers in $w_n$, by means of the properties of the support set. Hence the hypothesis $H_{n}'' = (H_{n}' \smallsetminus C') \cup R_{C'}$ is complete \& consistent w.r.t. $\mathit{\mathcal{E}_{tested}}$. The same argument shows that if $R_{C'}$ is further refined later on (case (iii) above), the resulting hypothesis remains complete an consistent w.r.t. $\mathit{\mathcal{E}_{tested}}$. Hence, when all windows have been tested, i.e. when $\mathcal{E}_{tested} = \mathcal{E}$, the resulting hypothesis $H_{n+1}$ is complete \& consistent w.r.t. $\mathcal{E} \cup w_n$ and furthermore, each window in $\mathcal{E}$ has been re-seen exactly once, thus $H_{n+1}$ is computed with a single pass over $\mathcal{E}$. 
\end{myproof}

\section{Discussion}
\label{sec:discussion}

Non-monotonic ILP, and XHAIL in particular, have some important properties, by means of which they extend traditional ILP systems. As briefly discussed in Section \ref{sec:learning_difficulties}, these properties are related to some challenging issues that occur when learning normal logic programs, which non-monotonic ILP addresses in a robust and elegant way. We next discuss which of these properties are preserved by ILED and which are sacrificed as a trade-off for efficiency, while briefly indicating directions for improvement in future work.

Like XHAIL, ILED aims for soundness, that is, hypotheses which cover all given examples. XHAIL ensures soundness by generalizing all examples in one go. In contrast, ILED preserves a memory of past experience for which newly acquired knowledge must account. Soundness imposes restrictions on the tasks on which ILED may be applied. In particular, we assume that the supervision is correct (i.e. it contains no contradictions or missing knowledge) and the domain is stationary, in the sense that knowledge already induced remains valid w.r.t. future instances, and retracting clauses or literals from the hypothesis at hand is never \emph{necessary} in order to account for new incoming example windows. ILED terminates in case its computations result in a dead-end, returning  no solution.
This results in treating cases such as \emph{concept drift} \citep{inthelex_concept_drift}, as noise. It is possible to relax the requirement for soundness and aim at an implementation that best-fits the training instances. Handling noise and concept drift are promising extensions of ILED.

XHAIL is a state-of-the-art system among its Inverse Entailment-based peer algorithms, in terms of completeness. That is, the hypotheses computable by XHAIL form a superset of those computable by other prominent Inverse Entailment systems like PROGOL and ALEPH \citep{xhail}. Although ILED preserves XHAIL's soundness, it does not preserve its completeness properties, due to the fact that ILED operates incrementally to gain efficiency. Thus there are cases where a hypothesis can be discovered by XHAIL, but be missed by ILED. As an example, consider cases where a target hypothesis captures long-term temporal relations in the data, as for instance, in the following clause:

\small
\begin{equation*}
\label{eq:no_locality_example}
\begin{array}{lr}
\myinitiatedAt(moving(X,Y),T) \leftarrow \\
\qquad \myhappens(walking(Y),T1),\\
\qquad T1<T.\\
\end{array}
\end{equation*}
\normalsize

\noindent In such cases, if the parts of the data that are connected via a long-range temporal relation are given in different windows, ILED has no way to correlate these parts in order to discover the temporal relation. However, one can always achieve XHAIL's functionality by increasing appropriately ILED's window size.  

An additional trade-off for efficiency is that not all of ILED's revisions are fully evaluated on the historical memory. For example, a new clause generated by a Kernel Set of an incoming window $w$ is selected randomly among a set of possible choices, which are equally good locally, i.e. in window $w$, but their quality may substantially differ globally. For instance, selecting a particular clause in order to cover a new example, may result in a large number of refinements and an unnecessarily lengthy hypothesis, as compared to one that may have been obtained by selecting a different initial clause. On the other hand, fully evaluating all possible choices throughout $\mathcal{E}$ requires extensive inference in $\mathcal{E}$. Thus simplicity and compression of hypotheses in ILED has been sacrificed for efficiency.

In ILED, a large part of the theorem proving effort that is involved in clause refinement reduces to computing subsumption between clauses, which is a hard task. Moreover, just as the historical memory grows over time, so do (in the general case) the support sets of the clauses in the running hypothesis, increasing the cost of computing subsumption. However, as in principle the largest part of a search space is redundant and the support set focuses only on its interesting parts, one would not expect that the support set will grow to a size that makes subsumption computation less efficient than inference over the entire $\mathcal{E}$. Moreover, the length of Kernel Set clauses (hence that of support clauses) is restricted by the size of incoming sliding windows. Smaller windows result to smaller clauses, making the computation of subsumption relations tractable. In addition, a number of optimization techniques have been developed over the years and several generic subsumption engines have been proposed \citep{django,resumer,subsumer}, some of which are able to efficiently compute subsumption relations between clauses comprising thousands of literals and hundreds of distinct variables. 

The basic idea behind ILED is to compress examples via Bottom Clause-like structures, in order to facilitate clause refinement, while learning a hypothesis incrementally. We see the idea behind the support set as being generic enough to be applied to any Inverse Entailment system that uses Bottom Clauses to guide the search, in order to provide support for more efficient clause refinement. In that case, the use of the support set should be modified accordingly to comply with the search method adopted by each system. For instance, in the work presented here, the support set works with XHAIL's search procedure, a minimality-driven, full search in the space of theories that subsume the Kernel Set, designed to address the non-monotonicity of normal logic programs. Different settings may be developed. For example, once the requirement for soundness is abandoned in an effort to address noise, a heuristic search strategy could be adopted, like for example PROGOL's $A^{*}$-like search. Different settings would require changes to the way the support set works. 

\section{Experimental evaluation}
\label{sec:experiments}
In this section, we present experimental results from two real-world applications: Activity recognition, using real data from the benchmark CAVIAR video surveillance dataset\footnote{\url{http://homepages.inf.ed.ac.uk/rbf/CAVIARDATA1/}}, as well as large volumes of synthetic CAVIAR data; and City Transport Management (CTM) using data from the PRONTO\footnote{\url{http://www.ict-pronto.org/}} project.   

Part of our experimental evaluation aims to compare ILED with XHAIL. To achieve this aim we had to implement XHAIL, because the original implementation was not publicly available until recently \citep{bragaglia2014}. 
All experiments were conducted on a 3.2 GHz Linux machine with 4 GB of RAM. The algorithms were implemented in Python, using the Clingo\footnote{\url{http://potassco.sourceforge.net/}} Answer Set Solver \citep{gebser2012answer} as the main reasoning component, and a Mongodb\footnote{\url{http://www.mongodb.org/}} NoSQL database for the historical memory of the examples. The code and datasets used in these experiments can be downloaded from \url{http://cer.iit.demokritos.gr/ILED/experiments}.

\subsection{Activity Recognition}
\label{sec:experiments_activity_recognition}

In activity recognition, our goal is to learn definitions of high-level events, such as \emph{fighting, moving} and \emph{meeting}, from streams of low-level events like \emph{walking, standing, active} and \emph{abrupt}, as well as spatio-temporal knowledge. We use the benchmark CAVIAR dataset for experimentation. Details on the CAVIAR dataset and more information about activity recognition applications 
may be found  in \citep{IJAIT}. Consider for instance the following definition of the \emph{fighting} high-level event:

\small
\noindent\begin{tabularx}{\textwidth}{@{}XXX@{}}
  \begin{equation}
    \label{eq:fighting_def_1}
    \begin{array}{lr}
    \myinitiatedAt(fighting(X,Y),T) \leftarrow \\
    \qquad \myhappens(active(X),T),\\
     \qquad \mynbf \  \myhappens(inactive(Y),T),\\
    \qquad \myholdsAt(close(X,Y,23),T).
   \end{array}
  \end{equation} & 
   \begin{equation}
    \label{eq:fighting_def_2} 
    \begin{array}{lr}
     \myinitiatedAt(fighting(X,Y),T) \leftarrow \\
     \qquad \myhappens(abrupt(X),T),\\
     \qquad \mynbf \  \myhappens(inactive(Y),T),\\    
     \qquad \myholdsAt(close(X,Y,23),T).
    \end{array}
  \end{equation}\\
  \begin{equation}
    \label{eq:fighting_def_3}
    \begin{array}{lr}
    \myterminatedAt(fighting(X,Y),T) \leftarrow \\
    \qquad \myhappens(walking(X),T), \\
    \qquad \mynbf \ \myholdsAt(close(X,Y,23),T).\\
    \end{array}
  \end{equation}&
  \begin{equation}
    \label{eq:fighting_def_4}
    \begin{array}{lr}
    \myterminatedAt(fighting(X,Y),T) \leftarrow \\
    \qquad \myhappens(running(X),T), \\
    \qquad \mynbf \ \myholdsAt(close(X,Y,23),T).\\
    \end{array}
  \end{equation}
\end{tabularx}

\normalsize

\noindent Clause (\ref{eq:fighting_def_1}) dictates that a period of time for which two persons $X$ and $Y$ are assumed to be fighting is initiated at time $T$ if one of these persons is \emph{active}, the other one is not \emph{inactive} and their distance is smaller than 23 pixel positions. Clause (\ref{eq:fighting_def_2}) states that  \emph{fighting} is initiated between two people when one of them moves \emph{abruptly}, the other is not \emph{inactive}, and the two persons are sufficiently close. Clauses (\ref{eq:fighting_def_3}) and (\ref{eq:fighting_def_4}) state that \emph{fighting} is terminated between two people when one of them walks or runs away from the other. 

 CAVIAR contains noisy data mainly due to human errors in the annotation \citep{list2005performance,IJAIT}. Thus, for the experiments we manually selected a noise-free subset of CAVIAR. The resulting dataset consists of 1000 examples (that is, data for 1000 distinct time points) concerning the high-level events \emph{moving}, \emph{meeting} and \emph{fighting}. These data, selected from different parts of the CAVIAR dataset, were combined into a continuous annotated stream of narrative atoms, with time ranging from 0 to 1000.
 
In addition to the real data, we generated synthetic data on the basis of the manually-developed CAVIAR event definitions  described in \citep{IJAIT}. In particular, streams of low-level events concerning four different persons were created randomly and were then classified using the rules of \citep{IJAIT}. The final dataset was obtained by generating negative supervision via the closed world assumption and appropriately pairing the supervision with the narrative. The generated data consists of approximately $10^5$ examples, which amounts to 100 MB of data.
 
The synthetic data is much more complex than the real CAVIAR data. This is due to two main reasons: First, the synthetic data includes significantly more initiations and terminations of a high-level event, thus much larger learning effort is required to explain it. Second, in the synthetic dataset more than one high-level event may be initiated or terminated at the same time point. This results in Kernel Sets with more clauses, which are hard to generalize simultaneously.

\subsubsection{ILED vs XHAIL}

The purpose of this experiment was to assess whether ILED can efficiently generate hypotheses comparable in size and predictive quality to those of XHAIL. To this end, we compared both systems on real and synthetic data using 10-fold cross validation with replacement. For the real data, 90\% of randomly selected examples, from the total of 1000 were used for training, while the remaining 10\% was retained for testing. At each run, the training data were presented to ILED in example windows of sizes 10, 50, 100. The data were presented in one batch to XHAIL. For the synthetic data, 1000 examples were randomly sampled at each run from the dataset for training, while the remaining data were retained for testing. Similar to the real data experiments, ILED operated on windows of sizes of 10, 50, 100 examples and XHAIL on a single batch.

Table \ref{table:expr_xhail_iled} presents the experimental results. 
Training times are significantly higher for XHAIL, due to the increased complexity of generalizing Kernel Sets that account for the whole set of the presented examples at once. These Kernel Sets consisted, on average, of 30 to 35 16-literal clauses, in the case of the real data, and 60 to 70 16-literal clauses in the case of the synthetic data. In contrast, ILED had to deal with much smaller Kernel Sets. The complexity of abductive search affects ILED as well, as the size of the input windows grows. ILED handles the learning task relatively well (in approximately 30 seconds) when the examples are presented in windows of 50 examples, but the training time increases almost 15 times if the window size is doubled. 

\newcommand{\ra}[1]{\renewcommand{\arraystretch}{#1}}
\begin{table}\centering
\scalebox{0.74}{
\begin{tabular}{@{}cccccccccc@{}}\toprule
& \multicolumn{3}{c}{ILED}  & \phantom{abc}& \multicolumn{1}{c}{XHAIL} \\
\cmidrule{2-4} \cmidrule{6-8} 
\textbf{Real CAVIAR data}  & $G = 10$ & $G = 50$ & $G = 100$  && $G = 900$ & ~ & ~ \\ \midrule

Training Time (sec)  & 34.15 ($\pm$ 6.87) & 23.04 ($\pm$ 13.50) & 286.74 ($\pm 98.87$)  && 1560.88 ($\pm 4.24$)\\
Revisions  & 11.2 ($\pm$ 3.05) & 9.1 ($\pm$ 0.32) & 5.2 ($\pm 2.1$)  && $-$ \\
Hypothesis size  & 17.82 ($\pm$ 2.18) & 17.54 ($\pm$ 1.5) & 17.5 ($\pm 1.43$)  &&  15 ($\pm 0.067$)  \\
Precision  & 98.713 ($\pm$ 0.052) & 99.767 ($\pm$ 0.038) & 99.971 ($\pm 0.041$)  && 99.973 ($\pm 0.028$)   \\
Recall  & 99.789 ($\pm$ 0.083) & 99.845 ($\pm$ 0.32) & 99.988 ($\pm 0.021$)  && 99.992 ($\pm 0.305$) &   \\
\hline \\
\textbf{Synthetic CAVIAR data} &  $G=10$ & $G=50$ & $G=100$ & ~ & $G = 1000$\\
\hline \\

Training Time (sec) & 38.92 ($\pm$ 9.15) & 33.87 ($\pm$ 9.74) & 468 ($\pm 102.62$)  && 21429 ($\pm 342.87$)\\
Revisions  & 28.7 ($\pm$ 9.34) & 15.4 ($\pm$ 7.5) & 12.2 ($\pm 6.23$)  && $-$ \\
Hypothesis size  & 143.52 ($\pm$ 19.14) & 138.46 ($\pm$ 22.7) & 126.43 ($\pm 15.8$)  && 118.18 ($\pm 14.48$)  \\
Precision  & 55.713 ($\pm$ 0.781) & 57.613 ($\pm$ 0.883) & 63.236 ($\pm 0.536$)  &&  63.822 ($\pm 0.733$)  \\
Recall  & 68.213 ($\pm$ 0.873) & 71.813 ($\pm$ 0.756) & 71.997 ($\pm 0.518$)  && 71.918 ($\pm 0.918$)  \\
\hline

\end{tabular}
}
\caption{\small Comparison of ILED and XHAIL. $G$ is the window granularity.}
\ra{1.5}
\label{table:expr_xhail_iled}
\end{table}
\normalsize

Concerning the size of the produced hypothesis, the results show that in the case of real CAVIAR data, the hypotheses constructed by ILED are comparable in size with a hypothesis constructed by XHAIL. In the case of synthetic data, the hypotheses returned by both XHAIL and ILED were significantly more complex. 
Note that for ILED the hypothesis size decreases as the window size increases. This is reflected in the number of revisions that ILED performs, which is significantly smaller when the input comes in larger batches of examples. 
In principle, the richer the input, the better the hypothesis that is initially acquired, and consequently, the less the need for revisions in response to new training instances. There is a trade-off between the window size (thus the complexity of the abductive search) and the number of revisions. A small number of revisions on complex data (i.e. larger windows) may have a greater total cost in terms of training time, as compared to a greater number of revisions on simpler data (i.e. smaller windows). For example, in the case of  window size 100 for the real CAVIAR data, ILED performs 5 revisions on average and requires significantly more time than in the case of a window size 50, where it performs 9 revisions on average. On the other hand, training times for windows of size 50 are slightly better than those obtained when the examples are presented in smaller windows of size 10. In this case, the ``unit cost'' of performing revisions w.r.t a single window are comparable between windows of size 10 and 50. Thus the overall cost in terms of training time is determined by the total number of revisions, which is greater in the case of window size  10.

Concerning predictive quality, the results indicate that ILED's precision and recall scores are comparable to those of XHAIL. 
For larger input windows, precision and recall are almost the same as those of XHAIL. This is because ILED produces better hypotheses from larger input windows. Precision and recall are smaller in the case of synthetic data for both systems, because the testing set in this case is much larger and complex than in the case of real data.

\subsubsection{ILED Scalability}

\begin{figure}[t]
\centering
\includegraphics[width=0.65\textwidth,height=0.25\textheight]{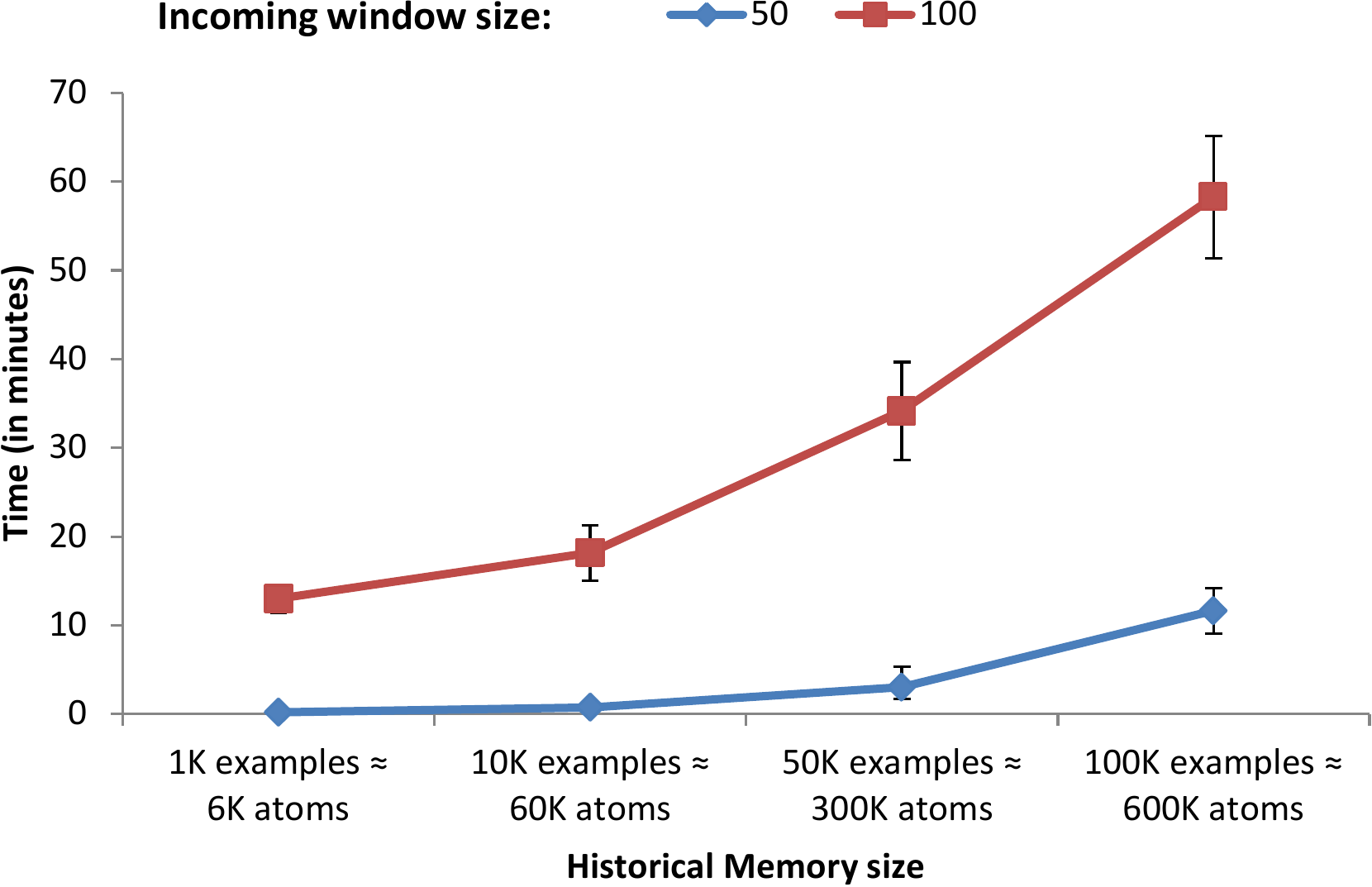}
\caption{\small Average times needed for ILED to revise an initial hypothesis in the face of new evidence presented in windows of size 10, 50 and 100 examples. The initial hypothesis was obtained from a training set of varying size (1K, 10K, 50K and 100K examples) which subsequently served as the historical memory.}
\label{fig:iled-scalability}

\end{figure}

The purpose of this experiment was to assess the scalability of ILED. The experimental setting was as follows: Sets of examples of varying sizes were randomly sampled from the synthetic dataset. Each such example set was used as a training set in order to acquire an initial hypothesis using ILED. Then a new window which did not satisfy the hypothesis at hand was randomly selected and presented to ILED, which subsequently revised the initial hypothesis in order to account for both the historical memory (the initial training set) and the new evidence. 
For historical memories ranging from $10^3$ to $10^5$ examples, a new training window of size 10, 50 and 100 was selected from the whole dataset. The process was repeated ten times for each different combination of historical memory and new window size. Figure \ref{fig:iled-scalability} presents the average revision times. The revision times for new window sizes of 10 and 50 examples are very close and therefore omitted to avoid clutter. The results indicate that revision time grows polynomially in the size of the historical memory.

\subsection{City Transport Management}

In this section we present experimental results from the domain of City Transport Management (CTM). We use data from the PRONTO\footnote{\url{http://www.ict-pronto.org/}} project. 
In PRONTO, the goal was to inform the decision-making
of transport officials by recognising high-level events related to  the punctuality of a public transport vehicle (bus or tram), passenger/driver comfort and safety. These high-level events were requested by the public transport control centre
of Helsinki, Finland, in order to support resource
management. Low-level events were provided by sensors installed in buses and trams, reporting on changes in position, acceleration/deceleration, in-vehicle temperature, noise level
and passenger density. At the time of the project,
the available datasets included only a subset of the
anticipated low-level event types as some low-level event detection components were not functional. For the needs of the project, therefore, a synthetic dataset was generated. The synthetic PRONTO data has proven to be considerably more challenging for event recognition than the real data, and therefore we chose the former for evaluating ILED \citep{artikisTKDE}. The CTM dataset contains $5\cdot 10^4$ examples, which amount approximately to 70 MB of data.

In contrast to the activity recognition application, the manually developed high-level event definitions of CTM that were used to produce the annotation for learning, form a hierarchy. In these hierarchical event definitions, it is possible to define a function level that maps all high-level events to non-negative integers as follows: A level-1 event is defined in terms of low-level events (input data) only. An level-$n$ event is defined in terms of at least one level-$n{-}1$ event and a possibly empty set of low-level events and high-level events of level below $n{-}1$. Hierarchical definitions are significantly more complex to learn as compared to non-hierarchical ones. This is  because initiations and terminations of events in the lower levels of the hierarchy appear in the bodies of event definitions in the higher levels of the hierarchy, hence all target definitions must be learnt simultaneously. As we show in the experiments, this has a striking effect on the required learning effort. A solution for simplifying the learning task is to utilize knowledge about the domain (the hierarchy), learn event definitions separately, and use the acquired theories from lower levels of the event hierarchy as non-revisable background knowledge when learning event definitions for the higher levels. Below is a fragment of the CTM event hierarchy:

\vspace*{-0.4cm}

\small
\begin{align} 
&\begin{array}{lr}
\label{eq:punctuality_init1}
    \myinitiatedAt(\mathit{punctuality(Id, \ nonPunctual), \ T}) \leftarrow \\
    \qquad \myhappens(stopEnter(Id, \ StopId, \ late), \ T).\\
    \end{array}\\
 &   \begin{array}{lr}
    \label{eq:punctuality_init2}
    \myinitiatedAt(\mathit{punctuality(Id, \ nonPunctual), \ T}) \leftarrow \\
    \qquad \myhappens(stopLeave(Id, \ StopId, \ early), \ T).\\
    \end{array}   \\ 
&\begin{array}{lr}
     \label{eq:punctuality_term1}
    \myterminatedAt(\mathit{punctuality(Id, \ nonPunctual), \ T}) \leftarrow \\
    \qquad \myhappens(stopEnter(Id, \ StopId, \ early), \ T).\\
    \end{array} \\  
    &\begin{array}{lr}
    \label{eq:punctuality_term2}
    \myterminatedAt(\mathit{punctuality(Id, \ nonPunctual), \ T}) \leftarrow \\
    \qquad \myhappens(stopEnter(Id, \ StopId, \ scheduled), \ T).\\
    \end{array}\\
& \begin{array}{lr}
    \label{eq:driving_qual_init1}
    \myinitiatedAt(\mathit{drivingQuality(Id, \ low), \ T}) \leftarrow \\
    \qquad \myinitiatedAt(punctuality(Id, \ nonPunctual), \ T),\\
    \qquad \myholdsAt(drivingStyle(Id, \ unsafe), \ T).\\
    \end{array}    \\
& \begin{array}{lr}
    \label{eq:driving_qual_init2}
    \myinitiatedAt(\mathit{drivingQuality(Id, \ low), \ T}) \leftarrow \\
    \qquad \myinitiatedAt(drivingStyle(Id, \ unsafe), \ T),\\
    \qquad \myholdsAt(punctuality(Id, \ nonPunctual), \ T).\\
    \end{array}    \\    
&\begin{array}{lr}
    \label{eq:driving_qual_term1}
    \myterminatedAt(\mathit{drivingQuality(Id, \ low), \ T}) \leftarrow \\
    \qquad \myterminatedAt(punctuality(Id, \ nonPunctual), \ T).\\
    \end{array}   \\
&\begin{array}{lr}
    \label{eq:driving_qual_term2}
    \myterminatedAt(\mathit{drivingQuality(Id, \ low), \ T}) \leftarrow \\     \qquad \myterminatedAt(drivingStyle(Id, \ unsafe), \ T).\\
    \end{array}     
\end{align}

\normalsize

\noindent Clauses (\ref{eq:punctuality_init1}) and (\ref{eq:punctuality_init2}) state that a period of time for which vehicle $\mathit{Id}$ is said to be \emph{non-punctual} is initiated if it enters a stop later, or leaves a stop earlier than the scheduled time. Clauses (\ref{eq:punctuality_term1}) and (\ref{eq:punctuality_term2}) state that the period for which vehicle $\mathit{Id}$ is said to be non-punctual is terminated when the vehicle arrives at a stop earlier than, or at the scheduled time. The definition of non-punctual vehicle uses two low-level events, $\mathit{stopEnter}$ and $\mathit{stopLeave}$. 

Clauses (\ref{eq:driving_qual_init1})-(\ref{eq:driving_qual_term2}) define \emph{low driving quality}. Essentially, driving quality is said to be low when the driving style is unsafe and the vehicle is non-punctual. Driving quality is defined in terms of high-level events (we omit the definition of driving style to save space). Therefore, the bodies of the clauses defining driving quality include \myinitiatedAt/2 and \myterminatedAt/2 literals.

\subsubsection{ILED vs XHAIL}

\begin{table}\centering
\scalebox{0.74}{
\begin{tabular}{@{}cccccccccc@{}}\toprule
& \multicolumn{2}{c}{ILED}  & \phantom{abc}& \multicolumn{1}{c}{XHAIL} \\
\cmidrule{2-4} \cmidrule{6-8} 
\textbf{~}  & $G = 5$ & $G = 10$   && $G = 20$  & ~ \\ \midrule

Training Time (\textbf{hours})  & 1.35 ($\pm$ 0.17) & 1.88 ($\pm$ 0.13) && 4.35 ($\pm 0.25$)\\
Hypothesis size  & 28.32 ($\pm$ 1.19) & 24.13 ($\pm$ 2.54)   &&  24.02 ($\pm 0.23$)  \\
Revisions  & 14.78 ($\pm$ 2.24) &  13.42 ($\pm$ 2.08)   && $-$ \\
Precision  &  63.344 ($\pm$ 5.24)  & 64.644 ($\pm$ 3.45)   &&  66.245 ($\pm$ 3.83)   \\
Recall  & 59.832 ($\pm$ 7.13) & 61.423 ($\pm$ 5.34)   && 62.567 ($\pm$ 4.65)    \\

\hline

\end{tabular}
}
\caption{\small Comparative performance of ILED and XHAIL on selected subsets of the CTM dataset each containing 20 examples. $G$ is the granularity of the windows. 
}
\ra{1.5}
\label{table:expr_ctm_zero_knowledge}
\end{table}
\normalsize 

In this experiment, we tried to learn simultaneously definitions for all target concepts, a total of nine interrelated high-level events, seven of which are level-1, one is level-2 and one is level-3. According to the employed language bias, each such high-level event must be learnt, while at the same time it may be present in the body of another high-level event in the form of (potentially negated) \myholdsAt/2, \myinitiatedAt/2, or \myterminatedAt/2 predicate. The total number of low-level events involved is 22.  

We used tenfold cross validation with replacement, on small amounts of data, due to the complexity of the learning task. In each run of the cross validation, we randomly sampled 20 examples from the CTM dataset, 90\% of which was used for training and 10\% was retained for testing. This example size was selected after experimentation, in order for XHAIL to be able to perform in an acceptable time frame. Each sample consisted of approximately 150 atoms (narrative and annotation). The examples were given to ILED in windows of granularity 5 and 10, and to XHAIL in one batch. Table \ref{table:expr_ctm_zero_knowledge} presents the average training times, hypothesis size, number of revisions, precision and recall.

ILED took on average 1-2 hours to complete the learning task, for windows of 5 and 10 examples, while XHAIL required more than 4 hours on average to learn hypotheses from batches of 20 examples. Compared to activity recognition, the learning setting requires larger Kernel Set structures that are hard to reason with. An average Kernel Set generated from a batch of just 20 examples consisted of approximately 30-35 clauses, with 60-70 literals each. 

Like the activity recognition experiments, precision and recall  scores for ILED are comparable to those of XHAIL, with the latter being slightly better. Unlike the activity recognition experiments, precision and recall had a large diversity between different runs. Due to the complexity of the CTM dataset, the constructed hypotheses had a large diversity, depending on the random samples that were used for training. For example, some high-level event definitions were unnecessarily lengthy and difficult to be understood by a human expert. On the other hand, some level-1 definitions could in some runs of the experiment, be learnt correctly even from a limited amount of data. Such definitions are fairly simple, consisting of one initiation and one termination rule, with one body literal in each case.

This experiment demonstrates several limitations of learning in large and complex applications. The complexity of the domain increases the intensity of the learning task, which in turn makes training times forbidding, even for small amount of data such as 20 examples (approximately 150 atoms). This forces one to process small sets of examples at at time, which in complex domains like CTM, results to over-fitted theories and rapid increase in hypothesis size. 

\subsubsection{Learning With Hierarchical Bias}
In an effort to improve the experimental results, we utilized domain knowledge about the event hierarchy in CTM and attempted to learn high-level events in different levels separately. To do so, we had to learn a complete definition from the entire dataset for a high-level event, before utilizing it as background knowledge in the learning process of a higher-level event. To facilitate the learning task further, we also used expert knowledge about the relation between specific low-level and high-level events, excluding from the language bias mode declarations which were irrelevant to the high-level event that is being learnt at each time. 

\begin{table}\centering
\scalebox{0.74}{
\begin{tabular}{@{}cccccccccc@{}}\toprule
& \multicolumn{3}{c}{ILED}  & \phantom{abc} \\
\cmidrule{2-4} \cmidrule{6-8} 
\textbf{level-1}  & $G = 10$ & $G = 50$ & $G = 100$   \\ \midrule

Training Time (min)  & 4.46 -- 4.88 & 5.78 -- 6.44 & 6.24 -- 6.88 \\
Revisions  & 2 -- 11 & 2 -- 9 & 2 -- 9 \\
Hypothesis size  & 4 -- 18 & 4 -- 16 & 4 -- 16    \\
Precision  & 100\% & 100\% & 100\%    \\
Recall  & 100\% & 100\% & 100\%   \\
\hline \\
\textbf{level-2} &  $G=10$ & $G=50$ & $G=100$ \\
\hline \\

Training Time (min)  & 8.76  & 9.14 & 9.86  \\
Revisions  & 24 & 17 & 17   \\
Hypothesis size  & 31 & 27 & 27   \\
Precision  & 100\% & 100\% & 100\%    \\
Recall  & 100\% & 100\% & 100\%    \\
\hline

\hline \\
\textbf{level-3} &  $G=10$ & $G=50$ & $G=100$ \\
\hline \\

Training Time (min)  & 5.78  & 6.14 & 6.78 \\
Revisions  & 6  & 5 & 5  \\
Hypothesis size  & 13 & 10 & 10    \\
Precision  & 100\% & 100\% & 100\%   \\
Recall  & 100\% & 100\% & 100\%    \\
\hline

\end{tabular}
}
\caption{\small ILED with hierarchical bias.}
\ra{1.5}
\label{table:expr_level_wise_learning}
\end{table}
\normalsize

The experimental setting was therefore as follows: Starting from the level-1 target events, we processed the whole CTM dataset in windows of 10, 50 and 100 examples with ILED. 
Each high-level event was learnt independently of the others. Once complete definitions for all level-1 high-level events were constructed, they were added to the background knowledge. Then we proceeded with learning the definition for the single  level-2 event. Finally, after  successfully constructing the level-2 definition, we performed learning in the top-level of the hierarchy, using the previously constructed level-1 and level-2 event definitions as background knowledge. We did not attempt a comparison with XHAIL, since due to the amounts of data in CTM, the latter is not able to operate on the entire dataset.

Table \ref{table:expr_level_wise_learning} presents the results. For level-1 events, scores are presented as minimum-maximum pairs. For instance, the training times for level-1 events with windows of 10 examples, ranges from 4.46 to 4.88 minutes. 
Levels 2 and 3 have just one definition each, therefore Table \ref{table:expr_level_wise_learning} presents the respective scores from each run. Training times, hypothesis sizes and overall numbers of revisions are comparable for all levels of the event hierarchy. Level-1 event definitions were the easiest to acquire, with training times ranging approximately between 4.50 to 7 minutes. This was expected since clauses in level-1 definitions are significantly simpler than level-2 and level-3 ones. 
The level-2 event definition was the hardest to construct with training times ranging between 8 and 10 minutes, while a significant number of revisions was required for all window granularities. The definition of this high-level event (\emph{drivingStyle}) is relatively complex, in contrast to the simpler level-3 definition, for which training times are comparable to the ones for level-1 events. 

The largest parts of training times were dedicated to checking an already correct definition against the part of the dataset that had not been processed yet. That is, for all target events, ILED converged to a complete definition relatively quickly, i.e. in approximately 1.5 to 3 minutes after the initiation of the learning process. From that point on, the extra time was spent on testing the hypothesis against the new incoming data. 

Window granularity slightly affects the produced hypothesis for all target high-level events. Indeed, the definitions constructed with windows of 10 examples are slightly larger than the ones constructed with larger window sizes of 50 and 100 examples. Notably, the definitions constructed with windows of granularity 50 and 100, were  found concise, meaningful and very close to the actual hand-crafted rules that were utilized in PRONTO. 



\section{Related work}
\label{sec:related}


A thorough review of the drawbacks of state-of-the-art ILP systems with respect to non-monotonic domains, as well as the deficiencies of existing approaches to learning Event Calculus programs can be found in \citep{xhail,sakama_2,nonmonotonic_inductive_logic_programming,otero2001induction,
otero2003induction}. The main obstacle, common to many learners which combine ILP with some form of abduction, like  PROGOL5 \citep{Muggleton_theory_compl}, ALECTO \citep{alecto}, HAIL \citep{hail} and IMPARO \citep{imparo}, is that they cannot perform abduction through negation and are thus essentially limited to Observational Predicate Learning. 

TAL \citep{tal} is a top-down non-monotonic learner which is able to solve the same class of problems as XHAIL. It obtains a top theory by appropriately mapping the ILP problem at hand to a corresponding ALP instance, so that solutions to the latter may be translated to solutions for the initial ILP problem. Recently, the main ideas behind TAL were employed in the ASPAL system \citep{corapi_asp_ilp}, an inductive learner which relies on Answer Set Programming as a unifying abductive-inductive framework. ASPAL obtains a top theory of \emph{skeleton rules} by forming all possible clause structures that may be formed from the mode declarations. Each such structure is complemented by a set of properly formed abducible predicates. Abductive reasoning on a proper \emph{meta-level} representation of the original ILP problem returns a set of such abducibles, which, due to their construction, allow to hypothesize on how variables and constants in the skeleton rules are linked together. Thus, the abduced atoms are prescriptions on how variable and constant terms in the original skeleton rules should be handled in order to obtain a hypothesis. This way, ASPAL may induce all possible hypotheses w.r.t to a certain ILP problem, as well as optimal ones, by computing minimal sets of abducibles.   

The combination of ILP with ALP has recently been applied to \linebreak \emph{meta-interpretive learning} (MIL), a learning framework, where the goal is to obtain hypotheses in the presence of a meta-interpreter. The latter is a higher-order program, hypothesizing about predicates or even rules of the domain. Given such background knowledge and sets of positive and negative examples, MIL uses abduction w.r.t. the meta-interpreter to construct first-order hypotheses. MIL can be realized both in Prolog and in Answer Set Programming, and it has been implemented in the METAGOL system \citep{muggleton2014meta}. Application examples involve learning definite clause grammars \citep{muggleton2014meta}, discovering relations between entities and learning simple robot action strategies \citep{muggleton2013meta}. MIL is an elegant framework, able to address difficult problems that are under-explored in traditional ILP, like handling predicate invention and learning mutually recursive programs. However, it has a number of important drawbacks. First, its expressivity is significantly limited, as IML is currently restricted to \emph{dyadic Datalog}, i.e. Datalog where the arity of each predicate is at most two. As noted in \citep{muggleton2014meta}, constructing meta-interpreters for richer fragments of first-order logic is not a straight-forward task and requires careful mathematical analysis. Second, given the increased computational complexity of higher-order reasoning, scaling to large volumes of data is a potential bottleneck for MIL.        

In the non-monotonic setting, traditional ILP approaches that cover the examples sequentially cannot ensure soundness and completeness \citep{sakama_2}. To deal with this issue, non-monotonic learners like XHAIL, TAL and ASPAL generalize all available examples in one go. The disadvantage of this approach, however, is poor scalability. A recent advancement which addresses the issue of scalability in non-monotonic ILP  is presented in \citep{raspal}. This approach combines the top-down, meta-level learning of TAL and ASPAL, with theory revision as ``non-monotonic ILP'' \citep{learning_rules_from_user_behaviour}, to address the ``grounding bottleneck'' in ASPAL's functionality. The top theory derived by ASPAL, as a starting point for its search, is based on combinations of the available mode declarations and grows exponentially with the length of its clauses. Thus, obtaining a ground program from this top theory is often very expensive and can cause a learning task to become intractable \citep{raspal}. RASPAL, the system proposed in \citep{raspal}, addresses this issue by imposing bounds on the length of the top theory. Partial hypotheses of specified clause length are iteratively obtained in a refinement loop. At each iteration of this loop, the partial hypothesis obtained from the previous refinement step is further refined using theory revision as described in \citep{learning_rules_from_user_behaviour}. The process continues until a complete and consistent hypothesis is obtained. The authors show that this approach results in shorter ground programs and derives a complete and consistent hypothesis, if one is derivable from the input data. An important difference between RASPAL and our approach is that the former addresses scalability as related to application domains, which may require a complex language bias, while our approach scales to potentially simpler, but massive volumes of sequential data, typical in temporal applications. 


TAL, ASPAL and RASPAL are top-down learners. In the work presented here, XHAIL, being a bottom-up non-monotonic learning system was the natural choice as the basis of our approach, since we intended to provide a clause refinement search bias by means of most-specific clauses, as in \citep{bot_clause_tr}. In that work, the Theory Revision system FORTE \citep{forte1} is enhanced by porting  PROGOL's bottom set construction routine to its functionality, towards a more efficient refinement operator. The resulting system, FORTE\_MBC, works as follows: When a clause $C$ must be refined, FORTE\_MBC uses mode declarations and an inverse entailment search in the background knowledge to construct a bottom clause from a positive example covered by $C$. It then searches for antecedents within the bottom clause. As in the case of ILED, the constrained search space results in a more efficient clause refinement process. However FORTE\_MBC (like FORTE itself) learns Horn theories and does not support non-Observational Predicate Learning, thus it cannot be used for the revision of Event Calculus programs. In addition, it cannot operate on an empty hypothesis (i.e. it cannot induce a hypothesis from scratch). Another important difference between FORTE\_MBC and ILED is the way that the former handles a potential incompleteness which may result from the specialization of a clause. In particular, once a clause is specialized, FORTE\_MBC checks again the whole database of examples. If some positive examples have become unprovable due to the specialization, FORTE\_MBC picks a different positive example covered by the initial, inconsistent clause $C$, constructs a new bottom clause and searches for a new specialization of clause $C$. The process continues until the original coverage in the example database is restored. In contrast, by means of the support set, the specializations performed by ILED preserve prior coverage in the historical memory, thus saving inference effort significantly.   

As mentioned in \citep{bot_clause_tr}, there is a renewed interest in scaling Theory Revision systems and applications in the last few years, due to the availability of large-scale domain knowledge in various scientific disciplines \citep{next_ten_years,chess_revision_muggleton_2010}. Temporal and stream data are no exception and there is a need for scalable Theory Revision techniques in event-based domains. However, most Theory Revision systems, such as the systems described in  \citep{forte,foil,audrey} limit their applicability to Horn theories. 

A well-known theory revision system is INTHELEX \citep{inthelex}. It is a fully incremental system that learns/revises Datalog theories and has been used in the study of several aspects of incremental learning. In particular, order effects in some simple learning tasks with ILP are discussed in \citep{esposito_backtracking,avoid_order_effects}, and concept drift in \citep{inthelex_concept_drift}. In \citep{inthelex_database} the authors present an approach towards scaling INTHELEX. In contrast to most ILP systems that keep all examples in the main memory, \citep{inthelex_database} follows an external memory implementation, which is the approach adopted by ILED. Moreover, in that work the authors associate clauses in the theory at hand with examples they cover, via a relational schema. Thus, when a clause is refined, only the examples that were previously covered by this clause are checked. Similarly, when a clause is generalized, only the negative examples are checked again. The scalable version of INTHELEX presented in  \citep{inthelex_database} maintains alternative versions of the hypothesis at each step, allowing to backtrack to previous states. In addition, it keeps in memory several statistics related to the examples that the system has already seen, such as the number of refinements that each example has caused, a ``refinement history'' of each clause, etc.

On the other hand, INTHELEX has some limitations that make it inappropriate for inducing/revising Event Calculus programs for event recognition applications. First, the restriction of its input language to Datalog limits its applicability to richer, relational event domains. For instance, complex relations between entities cannot be easily expressed in INTHELEX. Second, the use of background knowledge is limited, excluding for instance auxiliary clauses that may be used for spatio-temporal reasoning during learning time. Third, although INTHELEX uses abduction for the completion of imperfect input data, it relies on Observational Predicate Learning, meaning that it is not able to reason with predicates which are not directly observable in the examples. Therefore it cannot be used for learning event definitions.

\section{Conclusions}
\label{sec:concl}

We presented an incremental ILP system, ILED, for machine learning knowledge bases for event recognition, in the form of Event Calculus theories. ILED combines techniques from non-monotonic ILP and in particular, the XHAIL algorithm, with theory revision. It acquires an initial hypothesis from the first available piece of data, and revises this hypothesis as new data arrive. Revisions account for all accumulated experience. The main contribution of ILED is that it scales-up XHAIL to large volumes of sequential data with a time-like structure, typical of event-based applications. By means of a compressive memory structure that supports clause refinement, ILED has a scalable, single-pass revision strategy, thanks to which the cost of theory revision grows as a tractable function of the perceived experience. In this work, ILED was evaluated on an activity recognition application and a transport management application. The results indicate that ILED is significantly more efficient than XHAIL, without compromising the quality of the generated hypothesis in terms of predictive accuracy and hypothesis size. Moreover, ILED scales adequately to large data volumes which XHAIL cannot handle. Future work concerns mechanisms for handling noise and concept drift.

\section*{Acknowledgements}

This work is partly funded by the EU project SPEEDD (FP7 619435). We would like to thank the reviewers of the Machine Learning Journal for their valuable comments on the first version of the paper.




\begin{thebibliography}{73}
\providecommand{\natexlab}[1]{#1}
\providecommand{\url}[1]{{#1}}
\providecommand{\urlprefix}{URL }
\expandafter\ifx\csname urlstyle\endcsname\relax
  \providecommand{\doi}[1]{DOI~\discretionary{}{}{}#1}\else
  \providecommand{\doi}{DOI~\discretionary{}{}{}\begingroup
  \urlstyle{rm}\Url}\fi
\providecommand{\eprint}[2][]{\url{#2}}

\bibitem[{Ade and Denecker(1995)}]{ailp}
Ade H, Denecker M (1995) {AILP}: Abductive inductive logic programming. In:
  Proceedings of the 14th International Joint Conference on Artificial
  Intelligence

\bibitem[{Alrajeh et~al(2009)Alrajeh, Kramer, Russo, and
  S.Uchitel}]{learning_operational_requirements_from_goal_models}
Alrajeh D, Kramer J, Russo A, SUchitel (2009) Learning perational requirements
  from goal models. In: 31st International Conference on Software Engineering

\bibitem[{Alrajeh et~al(2010)Alrajeh, Kramer, Russo, and
  Uchitel}]{deriving_nonzeno_behaviour_models_from_goal_models_using_ILP}
Alrajeh D, Kramer J, Russo A, Uchitel S (2010) Deriving non-zeno behaviour
  models from goal models using ilp. Formal Aspects of Computing
  22(3-4):217--241

\bibitem[{Alrajeh et~al(2011)Alrajeh, Kramer, Russo, and
  Uchitel}]{an_inductive_approach_for_modal_transition_system_refinement}
Alrajeh D, Kramer J, Russo A, Uchitel S (2011) An inductive approach for modal
  transition system refinement. In: International Conference on Logic
  Programming

\bibitem[{Alrajeh et~al(2012)Alrajeh, Kramer, Russo, and
  S.Uchitel}]{learning_from_vacuously_satisfiable_scenariobased_specifications}
Alrajeh D, Kramer J, Russo A, SUchitel (2012) Learning from vacuously
  satisfiable scenario-based specifications. In: 15th International Conference
  on Fundamental Approaches to Software Engineering (FASE)

\bibitem[{Artikis et~al(2010)Artikis, Skarlatidis, and Paliouras}]{IJAIT}
Artikis A, Skarlatidis A, Paliouras G (2010) Behaviour recognition from video
  content: A logic programming approach. International Journal on Artificial
  Intelligence Tools 19(2):193--209

\bibitem[{Artikis et~al(2012)Artikis, Skarlatidis, Portet, and
  Paliouras}]{artikis2012logic}
Artikis A, Skarlatidis A, Portet F, Paliouras G (2012) Logic-based event
  recognition. The Knowledge Engineering Review 27(04):469--506

\bibitem[{Artikis et~al(2014)Artikis, Sergot, and Paliouras}]{artikisTKDE}
Artikis A, Sergot M, Paliouras G (2014) An event calculus for event
  recognition. IEEE Transactions on Knowledge and Data Engineering (TKDE)

\bibitem[{Athakravi et~al(2013)Athakravi, Corapi, Broda, and Russo}]{raspal}
Athakravi D, Corapi D, Broda K, Russo A (2013) Learning through hypothesis
  refinement using answer set programming. In: Proceedings of the 23rd
  International Conference of Inductive Logic Programming (ILP 2013)

\bibitem[{Badea(2001)}]{badea_2001}
Badea L (2001) A refinement operator for theories. In: Inductive Logic
  Programming, Springer, pp 1--14

\bibitem[{Biba et~al(2008)Biba, Basile, Ferilli, and
  Esposito}]{inthelex_database}
Biba M, Basile TMA, Ferilli S, Esposito F (2008) Improving scalability in ilp
  incremental systems

\bibitem[{Bragaglia and Ray(2014)}]{bragaglia2014}
Bragaglia S, Ray O (2014) Nonmonotonic learning in large biological networks.
  In: Proc. 24th Int. Conf. on Inductive Logic Programming

\bibitem[{Cattafi et~al(2010)Cattafi, Lamma, Riguzzi, and
  Storari}]{cattafi2010incremental}
Cattafi M, Lamma E, Riguzzi F, Storari S (2010) Incremental declarative process
  mining. Smart Information and Knowledge Management pp 103--127

\bibitem[{Cervesato and Montanari(2000)}]{cervesato2000calculus}
Cervesato I, Montanari A (2000) A calculus of macro-events: Progress report.
  In: Temporal Representation and Reasoning, 2000. TIME 2000. Proceedings.
  Seventh International Workshop on, IEEE, pp 47--58

\bibitem[{Chaudet(2006)}]{chaudet2006extending}
Chaudet H (2006) Extending the event calculus for tracking epidemic spread.
  Artificial Intelligence in Medicine 38(2):137--156

\bibitem[{Corapi et~al(2008)Corapi, Ray, Russo, Bandara, and
  Lupu}]{learning_rules_from_user_behaviour}
Corapi D, Ray O, Russo A, Bandara A, Lupu E (2008) Learning rules from user
  behaviour. In: Second International Workshop on the Induction of Process
  Models

\bibitem[{Corapi et~al(2010)Corapi, Russo, and Lupu}]{tal}
Corapi D, Russo A, Lupu EC (2010) Inductive logic programming as abductive
  search. In: Technical Communications of the 26th International Conference on
  Logic Programming ({ICLP})

\bibitem[{Corapi et~al(2011{\natexlab{a}})Corapi, De~Vos, Padget, Russo, and
  Satoh}]{corapi2011norm}
Corapi D, De~Vos M, Padget J, Russo A, Satoh K (2011{\natexlab{a}}) Norm
  refinement and design through inductive learning. In: Coordination,
  Organizations, Institutions, and Norms in Agent Systems VI, Springer, pp
  77--94

\bibitem[{Corapi et~al(2011{\natexlab{b}})Corapi, Russo, and
  Lupu}]{corapi_asp_ilp}
Corapi D, Russo A, Lupu E (2011{\natexlab{b}}) Inductive logic programming in
  answer set programming. In: ILP

\bibitem[{Denecker and Kakas(2002)}]{alp_3}
Denecker M, Kakas A (2002) Abduction in logic programming. Computational Logic:
  Logic Programming and Beyond 2407:402--437

\bibitem[{Dietterich et~al(2008)Dietterich, Domingos, Getoor, Muggleton, and
  Tadepalli}]{next_ten_years}
Dietterich TG, Domingos P, Getoor L, Muggleton S, Tadepalli P (2008) Structured
  machine learning: the next ten years. Machine Learning 73:3--23

\bibitem[{Duboc et~al(2009)Duboc, Paes, and Zaverucha}]{bot_clause_tr}
Duboc AL, Paes A, Zaverucha G (2009) Using the bottom clause and mode
  declarations in {FOL} theory revision from examples. Machine Learning
  76(1):73--107

\bibitem[{D{\v{z}}eroski(2010)}]{dvzeroski2010relational}
D{\v{z}}eroski S (2010) Relational data mining. Springer

\bibitem[{Eshghi and Kowalski(1989)}]{naf_abduction}
Eshghi K, Kowalski R (1989) Abduction compared with negation by failure. In:
  6th International Conference on Logic Programming

\bibitem[{Esposito et~al(2000)Esposito, Semeraro, Fanizzi, and
  Ferilli}]{inthelex}
Esposito F, Semeraro G, Fanizzi N, Ferilli S (2000) Multistrategy theory
  revision: Induction and abduction in inthelex. Machine Learning
  28(1-2):133--156

\bibitem[{Esposito et~al(2004)Esposito, Ferilli, Fanizzi, Basile, and
  Mauro}]{inthelex_concept_drift}
Esposito F, Ferilli S, Fanizzi N, Basile TMA, Mauro ND (2004) Incremental
  learning and concept drift in inthelex. Intelligent Data Analysis
  8(3):213--237

\bibitem[{Etzion and Niblett(2010)}]{etzion2010event}
Etzion O, Niblett P (2010) Event processing in action. Manning Publications Co.

\bibitem[{Fogel and Zaverucha(1998)}]{fogel1998normal}
Fogel L, Zaverucha G (1998) Normal programs and multiple predicate learning.
  In: Inductive Logic Programming, Springer, pp 175--184

\bibitem[{Gebser et~al(2012)Gebser, Kaminski, Kaufmann, and
  Schaub}]{gebser2012answer}
Gebser M, Kaminski R, Kaufmann B, Schaub T (2012) Answer set solving in
  practice. Synthesis Lectures on Artificial Intelligence and Machine Learning
  6(3):1--238

\bibitem[{Gelfond and Lifschitz(1988)}]{stable_model_semantics}
Gelfond M, Lifschitz V (1988) The stable model semantics for logic programming.
  In: International Conference on Logic Programming, pp 1070--1080

\bibitem[{Kakas and Mancarella(1990)}]{alp_1}
Kakas A, Mancarella P (1990) Generalised stable models: A semantics for
  abduction. In: ninth European Conference on Artificial Intelligence
  ({ECAI-90}), pp 385--391

\bibitem[{Kakas et~al(1993)Kakas, Kowalski, and Toni}]{alp_2}
Kakas A, Kowalski R, Toni F (1993) Abductive logic programming. Journal of
  Logic and Computation 2:719--770

\bibitem[{Kimber et~al(2009)Kimber, Broda, and Russo}]{imparo}
Kimber T, Broda K, Russo A (2009) Induction on failure: Learning connected horn
  theories. Logic Programming and Nonmonotonic Reasoning, Lecture Notes in
  Computer Science 5753:169--181

\bibitem[{Kowalski and Sergot(1986)}]{event_calculus}
Kowalski R, Sergot M (1986) A logic-based calculus of events. New Generation
  Computing 4(1):67–96

\bibitem[{Kuzelka and Zelezny(2008)}]{resumer}
Kuzelka O, Zelezny F (2008) A restarted strategy for eﬃcient subsumption
  testing. Fundamenta Informaticae 89

\bibitem[{Langley(1995)}]{langley_incremental_learning}
Langley P (1995) Learning in Humans and Machines: Towards an Interdisciplinary
  Learning Science, Elsevier, chap Order Effects in Incremental Learning

\bibitem[{Lavrac and Dzeroski(1993)}]{ilp_book}
Lavrac N, Dzeroski S (1993) Inductive Logic Programming: Techniques and
  Applications. Routledge

\bibitem[{Li and Lee(2009)}]{li2009mining}
Li HF, Lee SY (2009) Mining frequent itemsets over data streams using efficient
  window sliding techniques. Expert Systems with Applications 36(2):1466--1477

\bibitem[{Li et~al(2004)Li, Lee, and Shan}]{li2004efficient}
Li HF, Lee SY, Shan MK (2004) An efficient algorithm for mining frequent
  itemsets over the entire history of data streams. In: Proc. of First
  International Workshop on Knowledge Discovery in Data Streams

\bibitem[{List et~al(2005)List, Bins, Vazquez, and
  Fisher}]{list2005performance}
List T, Bins J, Vazquez J, Fisher RB (2005) Performance evaluating the
  evaluator. In: Visual Surveillance and Performance Evaluation of Tracking and
  Surveillance, 2005. 2nd Joint IEEE International Workshop on, IEEE, pp
  129--136

\bibitem[{Lloyd(1987)}]{lp_foundations_lloyd}
Lloyd J (1987) Foundations of Logic Programming. Springer

\bibitem[{Luckham(2001)}]{power_of_events}
Luckham D (2001) The Power of Events: An Introduction to Complex Event
  Processing in Distributed Enterprise Systems. Addison-Wesley Longman
  Publishing Co., Inc

\bibitem[{Luckham and Schulte(2008)}]{luckham_event_glossary}
Luckham D, Schulte R (2008) Event processing glossary version 1.1. Event
  Processing Technical Society

\bibitem[{Maggi et~al(2011)Maggi, Corapi, Russo, Lupu, and
  Visaggio}]{maggi2011revising}
Maggi FM, Corapi D, Russo A, Lupu E, Visaggio G (2011) Revising process models
  through inductive learning. In: Business Process Management Workshops,
  Springer, pp 182--193

\bibitem[{Maloberti and Sebag(2004)}]{django}
Maloberti J, Sebag M (2004) Fast theta-subsumption with constraint satisfaction
  algorithms. Machine Learning 55

\bibitem[{Mauro et~al(2004)Mauro, Esposito, Ferilli, and
  Basile}]{esposito_backtracking}
Mauro ND, Esposito F, Ferilli S, Basile TM (2004) A backtracking strategy for
  order-independent incremental learning. In: Proceedings of ECAI04

\bibitem[{Mauro et~al(2005)Mauro, Esposito, Ferilli, and
  110-121.}]{avoid_order_effects}
Mauro ND, Esposito F, Ferilli S, 110-121 TB (2005) Avoiding order effects in
  incremental learning. In: AIIA 2005: Advances in Artificial Intelligence,

\bibitem[{Moyle(2003)}]{alecto}
Moyle S (2003) An investigation into theory completion techniques in inductive
  logic. PhD thesis, University of Oxford

\bibitem[{Mueller(2006)}]{commonsense_reasoning}
Mueller E (2006) Commonsense Reasoning. Morgan Kaufmann

\bibitem[{Mueller(2008)}]{mueller_ec}
Mueller E (2008) Event calculus. Handbook of Knowledge Representation 3 of
  FAI:671--708

\bibitem[{Muggleton(1995)}]{IE_and_progol}
Muggleton S (1995) Inverse entailment and progol. New Generation Computing
  13(3\&4):245--286

\bibitem[{Muggleton and Bryant(2000)}]{Muggleton_theory_compl}
Muggleton S, Bryant C (2000) Theory completion using inverse entailment. In:
  International Conference on Inductive Logic Programming, pp 130--146

\bibitem[{Muggleton and Lin(2013)}]{muggleton2013meta}
Muggleton S, Lin D (2013) Meta-interpretive learning of higher-order dyadic
  datalog: Predicate invention revisited. In: Proceedings of the Twenty-Third
  international joint conference on Artificial Intelligence, AAAI Press, pp
  1551--1557

\bibitem[{Muggleton and Raedt(1994)}]{ilp_theory_and_methods}
Muggleton S, Raedt LD (1994) Inductive logic programming: Theory and methods.
  Journal of Logic Programming 19/20:629–679

\bibitem[{Muggleton et~al(2012{\natexlab{a}})Muggleton, Paes, Costa, and
  Zaverucha}]{chess_revision_muggleton_2010}
Muggleton S, Paes A, Costa VS, Zaverucha G (2012{\natexlab{a}}) Chess revision:
  acquiring the rules of chess variants through fol theory revision from
  examples. In: Inductive Logic Programming

\bibitem[{Muggleton et~al(2012{\natexlab{b}})Muggleton, Raedt, Poole, Bratko,
  Flach, Inoue, and Srinivasan}]{ilp_turns_20}
Muggleton S, Raedt LD, Poole D, Bratko I, Flach P, Inoue K, Srinivasan A
  (2012{\natexlab{b}}) {ILP} turns 20 - biography and future challenges.
  Machine Learning 86(1):3--23

\bibitem[{Muggleton et~al(2014)Muggleton, Lin, Pahlavi, and
  Tamaddoni-Nezhad}]{muggleton2014meta}
Muggleton SH, Lin D, Pahlavi N, Tamaddoni-Nezhad A (2014) Meta-interpretive
  learning: application to grammatical inference. Machine Learning 94(1):25--49

\bibitem[{Otero(2001)}]{otero2001induction}
Otero RP (2001) Induction of stable models. In: Inductive Logic Programming,
  Springer, pp 193--205

\bibitem[{Otero(2003)}]{otero2003induction}
Otero RP (2003) Induction of the effects of actions by monotonic methods. In:
  Inductive Logic Programming, Springer, pp 299--310

\bibitem[{Paschke(2005)}]{paschke}
Paschke A (2005) Eca-ruleml: An approach combining eca rules with temporal
  interval-based kr event logics and transactional update logics. Tech. rep.,
  Technische Universitat Munchen

\bibitem[{Quinlan(1990)}]{foil}
Quinlan JR (1990) Learning logical definitions from relations. Machine Learning
  5:239–266

\bibitem[{Ray(2006)}]{ray_abduction_for_induction}
Ray O (2006) Using abduction for induction of normal logic programs. In:
  ECAI’06 Workshop on Abduction and Induction in Artiﬁcial Intelligence and
  Scientiﬁc Modelling

\bibitem[{Ray(2009)}]{xhail}
Ray O (2009) Nonmonotonic abductive inductive learning. Journal of Applied
  Logic 7(3):329--340

\bibitem[{Ray et~al(2003)Ray, Broda, and Russo}]{hail}
Ray O, Broda K, Russo A (2003) Hybrid abductive inductive learning: A
  generalisation of progol. In: International Conference in Inductive Logic
  Programming ({ILP}), pp 311--328

\bibitem[{Richards and Mooney.(1995)}]{forte1}
Richards B, Mooney R (1995) Automated refinement of first-order horn clause
  domain theories. Machine Learning 19(2):95--131

\bibitem[{Richards and Mooney(1991)}]{forte}
Richards BL, Mooney RJ (1991) First order theory revision. In: 8th
  International Workshop on Machine Learning, p 447–451

\bibitem[{Sakama(2000)}]{sakama_ie_naf}
Sakama C (2000) Inverse entailment in nonmonotonic logic programs. In: n
  Proceedings of the 10th International Conference on Inductive Logic
  Programming

\bibitem[{Sakama(2001)}]{nonmonotonic_inductive_logic_programming}
Sakama C (2001) Non-monotonic inductive logic programming. In: Logic
  Programming and Non-Monotonic Reasoning

\bibitem[{Sakama(2005)}]{sakama_2}
Sakama C (2005) Induction from answer sets in nonmonotonic logic programs. ACM
  Transactions on Computational Logic 6 (2):203–231

\bibitem[{Santos and Muggleton(2010)}]{subsumer}
Santos J, Muggleton S (2010) Subsumer: A prolog theta-subsumption engine. In:
  Technical Communications of the 26th International Conference on Logic
  Programming, Leibniz International Proceedings in Informatics

\bibitem[{Sloman and Lupu(2010)}]{engineering_policybased_ubiquitous_systems}
Sloman M, Lupu E (2010) Engineering policy-based ubiquitous systems. The
  Computer Journal 53(5):1113--1127

\bibitem[{Wogulis and Pazzani(1993)}]{audrey}
Wogulis J, Pazzani M (1993) A methodology for evaluating theory revision
  systems: Results with audrey ii. In: 13th Interantional Joint Conference in
  Artificial Intelligence {IJCAI}, pp 1128--1134

\bibitem[{Wrobel(1996)}]{wrobel}
Wrobel S (1996) First order theory refinement. In: Raedt LD (ed) Advances in
  Inductive Logic Programming, pp 14 -- 33

\end{thebibliography}


\begin{appendices}

\section{Notions from (Inductive) Logic Programming} 
\label{app:appendix_lp}

\begin{definition}[\textbf{Basic notions from Logic Programming} \citep{lp_foundations_lloyd}]
\label{def:LP_basic_notions}
A term is a constant, a variable, or an expression of the form $f(a_1,\ldots,a_n)$ where $f$ is a function symbol and $a_1,\ldots,a_n$ are terms. A term substitution is a function from the set of terms to itself. An atom is an expression of the form $p(a_1,\ldots,a_n)$ where $p$ is a predicate symbol and $a_1,\ldots,a_n$ are terms. A literal is either an atom $a$ (positive literal) or its negation $\mynbf \ a$ (negative literal). A clause $C$ is an expression of the form $a\leftarrow b_1,\ldots,b_n$ where $a$ is an atom and $b_1,\ldots,b_n$ are literals. $a$ is called the head of clause $C$, and $\{b_1,\ldots,b_n\}$ is called the body of the clause. A fact is a clause of the form $\mathit{a\leftarrow true}$ and an integrity constraint is a clause of the form $\mathit{false \leftarrow a}$. A logic program is a collection of clauses. A clause or a logic program is \emph{Horn} if it contains no negated literals and \emph{normal} otherwise.
\end{definition}

\begin{definition}[\textbf{Interpretations and models} \citep{stable_model_semantics}]
\label{def:interpretations_and_models} 
Given a logic program $\Pi$ an Herbrand interpretation $I$ is a subset of the set of all possible groundings of $\Pi$. $I$ satisfies a literal $a$ (resp. \mynbf$ \ a$) iff $a \in I$ (resp. $a \notin I$). $I$ satisfies a set of ground atoms iff it satisfies each one of them and it satisfies a ground clause iff it satisfies the head, or does not satisfy at least one body literal. $I$ is a Herbrand model of $\Pi$ iff it satisfies every ground instance of every clause in $\Pi$ and it is a minimal model iff no strict subset of $I$ is a model of $\Pi$. $I$ is a \emph{stable model} of $\Pi$ iff it is a minimal model of the Horn program that results from the ground instances of $\Pi$ after the removal of all clauses with a negated literal not satisfied by $I$, and  all negative literals from the remaining clauses. 
\end{definition}

\begin{definition}[\textbf{Mode Declarations} \citep{IE_and_progol}]
\label{def:mode_declarations}
A \emph{mode declaration} is either a head or body declaration, respectively, $\mathit{modeh(s)}$ and $\mathit{modeb(s)}$, where $s$ is called a schema. A schema $s$ is a ground literal containing \emph{placemarkers}. A placemarker is either $\mathit{+type}$ (input) $\mathit{-type}$ (output) or $\mathit{\#type}$ (ground), where $\mathit{type}$ is a constant. The distinction between input and output terms in mode declarations is that any input term in a body literal must be an input term in the head, or an output term in some preceding body literal. 
\end{definition}

\begin{definition}[\textbf{Mode language} \citep{IE_and_progol}]
\label{def:mode_language}
A set $M$ of mode declarations defines a language $\mathcal{L}(M)$. A clause $C$ is in $\mathcal{L}(M)$ if it results from the declarations in $M$ by replacing input and output placemarkers by variables and replacing ground placemarkers with ground terms. In particular $C \in \mathcal{L}(M)$ iff its head atom (respectively each of its body literals) is constructed from the schema $s$ in a $\mathit{modeh(s)}$ atom (resp. in a $\mathit{modeb(s)}$ atom) in $M$ as follows:
\begin{itemize}
\item By replacing an output $(-)$ placemarker by a new variable.
\item By replacing  an input $(+)$ placemarker by a variable that appears in the head atom, or in a previous body literal.
\item By replacing a ground $(\#)$ placemarker by a ground term.
\end{itemize}
A hypothesis $H$ is in $\mathcal{L}(M)$ iff $C\in \mathcal{L}(M)$ for each $C\in H$.
\end{definition}

\begin{definition}[\textbf{Variable depth} \citep{IE_and_progol}]
\label{def:variable_depth}
Let $C$ be a clause and $X$ a variable symbol. The depth $d(X)$ of $X$ is defined recursively as follows:\\

$
d(X) =
\left\{
	\begin{array}{ll}
		0  & \mbox{if } X \mbox{is in the head of } C\\
		(min_{Y\in V_{X}}d(Y)) + 1 & \mbox{else} \\
	\end{array}
\right.
$\\ \\
where $V_{X}$ are the variable symbols that appear in all literals in the body of $C$ in which $X$ also appears.
\end{definition}

\begin{definition}[\textbf{Depth-bound mode language} \citep{IE_and_progol}]
\label{def:depth_mode_language}
Let $M$ be a set of mode declarations, $i$ a non-negative integer and $C$ a clause. $C$ is in the depth-bounded mode language $\mathcal{L}_i(M)$ iff $C \in \mathcal{L}(M)$ (see Definition \ref{def:mode_language}) and for each variable symbol $X$ that appears in $C$ it holds that $d(X) \leq i$ (see Definition \ref{def:variable_depth}). A hypothesis $H$ is in $\mathcal{L}_i(M)$ iff $C\in \mathcal{L}_i(M)$ for each $C\in H$.
\end{definition}

\begin{definition}[\textbf{Most-specific clause relative to a set of examples}]
\label{def:most_specific_clause}
Let $\mathcal{L}_i(M)$ be the depth-bounded mode language as in Definition \ref{def:depth_mode_language}, $E$ a set of examples and $B$ some background theory. Let $\mathcal{L}_i(M,E) = \{C \in \mathcal{L}_i(M) \ | \ B\cup C \vDash E\}$. A clause  $\bot \in \mathcal{L}_i(M,E)$ is most-specific, relative to $E$, iff it does not $\theta$-subsume any other clause in $\mathcal{L}_i(M,E)$.
\end{definition}

\end{appendices}

\end{document}